\documentclass[journal]{IEEEtran}
\IEEEoverridecommandlockouts

\usepackage{cite}
\usepackage{amsmath,amssymb,amsfonts}
\usepackage{multirow}
\usepackage{threeparttable}
\usepackage{float}
\usepackage{graphicx}
\usepackage{textcomp}
\usepackage{stfloats}
\usepackage{xcolor}
\usepackage{booktabs} 
\usepackage{array}
\usepackage{algorithm}
\usepackage{algpseudocode}
\usepackage{subcaption}
\usepackage[justification=centering]{caption}
\usepackage{pifont}  
\usepackage{orcidlink}
\newcommand{\comment}[1]{}

\newcommand{\orcidicon}[1]{\href{https://orcid.org/#1}{\textcolor{black}{\aiOrcid}}}
\hypersetup{hidelinks} 

\ifCLASSINFOpdf
\else
\fi
\begin{document}
%
\title{Adaptive Residual Transformation for Enhanced Feature-Based OOD Detection in SAR Imagery}
%
%
%

\author{Kyung-Hwan Lee \orcidlink{0000-0002-4832-0420}
        and~Kyung-Tae Kim,~\IEEEmembership{Member, IEEE}

\thanks{Kyung-Hwan Lee is with the Next Generation Defence Multidisciplinary
Technology Research Center, Pohang University of Science and Technology,
Pohang 790-784, South Korea (e-mail: kyunghwanlee@postech.ac.kr) \\
\hspace*{1.5em}Kyung-Tae Kim is with the Department of Electrical Engineering, Pohang University of Science and Technology, Pohang 790-784, South Korea (e-mail: kkt@postech.ac.kr) \\
Corresponding Author: Kyung-Tae Kim
}}

%
%

\markboth{IEEE Transactions on Geoscience and Remote Sensing,~Vol.~X, No.~X, Month~Year}%
{Lee \MakeLowercase{\textit{et al.}}: Adaptive Residual Transformation for Enhanced Feature-Based OOD Detection in SAR Imagery}
%



\maketitle

\begin{abstract}
Recent advances in deep learning architectures have enabled efficient and accurate classification of pre-trained targets in Synthetic Aperture Radar (SAR) images. Nevertheless, the presence of unknown targets in real battlefield scenarios is unavoidable, resulting in misclassification and reducing the accuracy of the classifier. Over the past decades, various feature-based out of distribution (OOD) approaches have been developed to address this issue, yet defining the decision boundary between known and unknown targets remains challenging. 
\comment{
Detection rates of unknown targets unstably fluctuate depending on the decision threshold and the type/number of known/unknown targets.}
Additionally, unlike optical images, detecting unknown targets in SAR imagery is further complicated by high speckle noise, the presence of clutter, and the inherent similarities in back-scattered microwave signals. In this work, we propose transforming feature-based OOD detection into a class-localized feature-residual-based approach, demonstrating that this method can improve stability across varying unknown targets' distribution conditions.
\comment{
Initially, we extracted feature vectors from the penultimate layer during the evaluation phase. Using these feature representations, we computed both in-class and inter-class standardized residuals within the known class labels. The unknown target detection model utilizes these two residual types to learn about each label's distributional characteristics. During the test phase, the penultimate layer feature vectors from the test data are paired with the known target feature vectors for each class label to generate the test residual samples. The test residuals with each known label are then input into the pre-trained model to discriminate between known and unknown targets.}
Transforming feature-based OOD detection into a residual-based framework offers a more robust reference space for distinguishing between in-distribution (ID) and OOD data, particularly within the unique characteristics of SAR imagery. This adaptive residual transformation method standardizes feature-based inputs into distributional representations, enhancing OOD detection in noisy, low-information images. Our approach demonstrates promising performance in real-world SAR scenarios, effectively adapting to the high levels of noise and clutter inherent in these environments. These findings highlight the practical relevance of residual-based OOD detection for SAR applications and suggest a foundation for further advancements in unknown target detection in complex, operational settings.
\end{abstract}

\begin{IEEEkeywords}
SAR (Synthetic Aperture Radar), ATR (Automatic Target Recognition), Standardization, Feature extraction, Penultimate layer, Anomaly Detection, ID (In Distribution), OOD (Out of Distribution), Distributional Inputs, Unknown Target Detection
\end{IEEEkeywords}

%
\IEEEpeerreviewmaketitle

\section{Introduction}
%
%
%
%

\IEEEPARstart{C}{onventional} synthetic aperture radar automatic target recognition (SAR-ATR) experiments typically rely on meticulously crafted SAR images, such as those in the Moving and Stationary Target Acquisition and Recognition (MSTAR) database \cite{b1}, achieving near-perfect detection and classification rates with state-of-the-art deep learning algorithms \cite{b2,b3,b4,b5}. However, their outstanding performance fails to translate to real-world applications due to the insufficient size, quality, and completeness of actual datasets \cite{b6}.
In addition to these dataset limitations, the test stage will inevitably encounter untrained targets and clutter in real-time, on-site scenarios. The presence of untrained targets poses a significant challenge, which is further complicated by the previously mentioned inadequacies of the trainable datasets. This combination makes detecting and classifying unknown targets and clutter even more difficult \cite{b7, b8}.

Multiple deep-learning techniques for detecting unknown targets for SAR images have been developed over the past decades. For example, Zero-Shot Learning (ZSL) addresses this by learning from known classes and inferring unknown class samples without overlapping training and test sets, involving the extraction of image and semantic feature vectors \cite{b9}. In SAR–ATR applications of ZSL, it is essential to establish a stable and comprehensible reference space. This ensures the identification of unknown targets accurately by incorporating the prior knowledge of known targets. This deep-learning approach might be ideal because the decision boundary is learned during the training phase and, hence, less sensitive than the user-defined threshold setup \cite{b10, b11, b12, b13, b14, b15, b16, b17}. However, to date, ZSL-based unknown target detectors have been shown to be ineffective in a robust setup of the reference space. As an alternative to resolve this issue, feature extraction from the deep learning network layer has been attempted with the joint discrimination of feature extraction network (FEN), Kullback–Leibler divergence (KLD), and relative position angle (RPA) \cite{b18}.

A modified polar mapping classifier (M-PMC) was also considered for unknown target detection on machine learning and pattern recognition \cite{b19}. The aforementioned methods are fairly efficient when the detector encounters a few unknown classes with sufficiently large train data sizes. However, they still face a severe and fundamental problem: the threshold for distinguishing known and unknown classes significantly varies with the number and type of unknown classes. Setting a threshold to discriminate between trained and unknown targets is necessary because their decision boundary has not been trained due to the unavailability of the complete dataset. The threshold value for a typical unknown target classifier is generally pre-tuned through experimentation with a few testing samples, depending on the decision metric variation. However, as the detector encounters more new unknown targets, this threshold becomes progressively less accurate and eventually loses its ability to filter out the unknown targets effectively \cite{b18, b19, b20, b21, b22, b23, b24}. In real-world battlefield scenarios, the number and type of unknown targets are not pre-determined; hence, we cannot continuously calibrate the threshold on a case-by-case basis.

\begin{table*}[htbp]
\centering
\caption{\textsc{Examples of Out-of-Distribution Methods in Computer Vision Area}}
\begin{tabular}{llllllll}
\hline
\hline
\multicolumn{1}{c}{\multirow{2}{*}{OOD Method}} & \multirow{2}{*}{Uses} & \multicolumn{3}{c}{Base} & \multicolumn{2}{c}{Type}             & \multirow{2}{*}{Ref.} \\ \cline{3-7}
\multicolumn{1}{c}{}                            &                       & Feature & Logits/ Prob. & Gradient & Post-processing & Internal Structure &                       \\ \hline
MSP                                             &                       &         & $\checkmark$  &  & $\checkmark$       &                    &  \cite{b22}                     \\ \hline
OpenMax                                         &               & $\checkmark$  & $\triangle$  &       &  $\checkmark$       &           & \cite{b26}                      \\ \hline
MaxLogit                                        &                       &         & $\checkmark$ &  & $\checkmark$       &              & \cite{b28}                      \\ \hline
Energy                                          &                       &  & $\checkmark$ &      & $\checkmark$       &                    & \cite{b29}                      \\ \hline
ReAct                                        &                       & $\checkmark$  &        &  & $\checkmark$       &           & \cite{b59}                      \\ \hline
ViM                                             &                       & $\checkmark$  & $\triangle$ &      & $\checkmark$      &                    & \cite{b31}                      \\ \hline
ODIN                                            &                       &         & $\checkmark$ & $\checkmark$ & $\checkmark$    &                    & \cite{b27}                      \\ \hline
Mahalanobis dist.                               &                       & $\checkmark$  &        & $\triangle$        & $\checkmark$       &              & \cite{b24}                      \\ \hline
VI/MC Dropout                                   &                       &   &  &      &                 & $\checkmark$             & \cite{b32, b33}                      \\ \hline
Deep Ensembles                                  &                       &   &        &  &              & $\checkmark$             & \cite{b34}                      \\ \hline
RPL                       &                       & $\checkmark$  &  &      & $\triangle$       & $\checkmark$                   & \cite{b41}            \\ \hline
\end{tabular}
\begin{tablenotes}
\footnotesize
\item Check signs: main usage of the corresponding feature or type, Triangle signs: Either (1) also used in addition to the space where the check sign is applied or (2) could potentially be extended for the development of detector with the corresponding base and the type
\item OOD: Out of Distribution, MD: Mahalanobis distance, MSP: Maximum Softmax Probability, ReAct: Rectified Activations, ViM: Virtual-logit Matching, ML: Maximum Logits, VI: Variational Inference, MC Dropout: Monte Carlo Dropout, ODIN: Out of Distribution Detector for Neural Networks, RPL: Reciprocal Point Learning
\end{tablenotes}
\end{table*}

Beyond attempts at detecting unknown targets in SAR imagery, multiple novel approaches have been developed for Out-of-Distribution (OOD) detection across broader fields, including computer vision, natural language processing, and automatic speech recognition \cite{b20, b21, b22, b23, b24, b25, b26, b27, b28, b29, b30, b31, b32, b33, b34, b35, b36, b37, b38, b39,b40, b41, b42}.
Anomaly detection methods can be categorized into internal network structure-based and post-processing-based approaches, as summarized in Table I. Internal network structure-based methods involve mainly altering the original classification neural networks or re-training of neural networks for OOD detection. In contrast, post-processing methods perform OOD detection mainly using the outputs of these networks, often without requiring changes to the network’s classification structure, though some exceptions, such as ODIN (Out of Distribution Detector for Neural Network) and RPL (Reciprocal Point Learning) \cite{b27} \cite{b41}, may involve structural modifications and re-training. Generally, internal structure-based methods cannot be applied to fully trained neural networks, as they require adjustments during the training process. Examples of OOD detection methods that involve internal network modifications include variational inference and Monte Carlo Dropout, which use Bayesian deep learning and predictive distributions derived from posterior evaluations; deep ensembles, which rely on statistical inference based on uncertainty distributions from multiple model predictions; and generative models/discriminative models with adversarial learning, which typically use reconstruction errors or likelihood estimation during re-training to distinguish OOD samples \cite{b30, b31, b32, b33, b34, b35, b36, b37, b38, b39, b40, b41, b42}. Particularly, approaches based on the Bayesian principle involve obtaining the posterior distribution by updating prior beliefs with observed data likelihoods, with the predictive distribution derived by integrating over this posterior distribution \cite{b30, b31, b32, b33, b36, b37}. Although Bayesian deep learning offers robust uncertainty estimates in network outputs, its high computational complexity and longer inference times can hinder practicality in real-world applications.

Conversely, post-processing methods extract information from feature, logit, probability (softmax), or gradient spaces and generally require less computation time. Logit/softmax-based OOD detection methods include Maximum Softmax Probability (MSP) \cite{b22}, MaxLogit \cite{b28}, Energy-based approaches \cite{b29}, and Out-of-Distribution Detector for Neural Networks (ODIN), which may involve additional network adjustments \cite{b27}. Feature-based methods include OpenMax, ReAct (Rectified Activations), ViM (Virtual Logits Matching), and Mahalanobis distance-based techniques \cite{b26, b59, b31, b24}. For example, in logit/softmax-based OOD detection methods, MSP determines OOD/ID status based on whether the highest softmax probability—representing the model’s confidence score—exceeds a threshold established from the training dataset distribution. In feature-based OOD detection methods, OpenMax extends softmax by incorporating open set recognition, adjusting probability estimates to include an 'unknown' class likelihood based on the distance of activations from known class centers \cite{b26}. The Mahalanobis distance method assumes that ID class features follow Gaussian distributions. It computes the Mahalanobis distance—a covariance-adjusted metric—between the test sample’s feature vector and each class's mean feature vector. A higher distance suggests a greater likelihood of being OOD, as it falls further from the learned in-distribution feature space \cite{b24}. ViM enhances the logit space by introducing a virtual logit, derived from projecting the test sample's feature vector onto principal components of training class feature vectors via Principal Component Analysis (PCA). This projection creates a distance metric between ID and OOD samples. By evaluating the proximity of a sample to this virtual logit, the algorithm can effectively distinguish OOD from ID samples \cite{b31}.

Feature-based OOD detection methods are particularly advantageous due to their enriched representations of target characteristics, allowing for extended variability in OOD detection algorithms. Nonetheless, attempts to apply these OOD detection methods in the SAR imagery domain are highly limited. Furthermore, the feasibility and comparable performance seen in optical image datasets are uncertain, as SAR images typically have lower resolution and higher noise levels than optical images. Regardless of the underlying approach, adapting OOD detection methods to suit SAR imagery is essential but has not been thoroughly investigated.
To address these challenges, our work introduces a reshaping approach that bypasses conventional feature-space OOD detection methods. Instead, it transforms feature information into localized feature-residuals by calculating within-class and between-class differences across feature vectors on a per-class basis. Through this approach, we modify the type and interpretation of inputs without altering the underlying feature-based OOD detection algorithm, thereby preserving generalizability for application across various feature-based OOD detection methods. Our contributions to unknown target detection in SAR imagery are as follows:
\begin{enumerate}
    \item This paper proposes a novel approach that leverage class-localized residuals of feature vectors, enabling generalizability to prior feature-based OOD detection algorithms. We detail the statistical formulation and the construction of a reference space within this framework. Performance metrics are evaluated for residual-based adaptations of prior OOD detection methods under varying unknown target conditions and compared to their feature-based counterparts, providing a comprehensive analysis of the suitability and effectiveness of residual-based approaches.

    \item Furthermore, we demonstrate that our approximate standardization techniques, based on residuals, effectively mitigate poor OOD detection performance in high-clutter, noisy, and low-information scenarios. This approach reduces threshold variability caused by untrained test target samples, enhancing detection accuracy. We then examine the advantages of our approach over original feature-based methods, particularly in the context of SAR image characteristics.
    
    \item Finally, we constrain advanced OOD detection algorithms aimed at capturing subtle distributional patterns and improving accuracy through a residual-based approach. We discuss how the intrinsic distributional characteristics of residuals can be analyzed for effective OOD detection, as in-class and inter-class residuals exhibit complex, distinct patterns. Additionally, we explore future directions for optimizing threshold algorithms within our residual-based framework.
\end{enumerate}

The remainder of this paper is organized as follows: In Section II, we address the challenges of detecting unknown targets in real battlefield scenarios and discuss the need for a new foundation in OOD detection algorithms, particularly in light of the differences between optical and SAR imagery. We introduce our approach as a potential solution to these challenges in SAR imagery. Section III outlines our methodology for class-localized feature-residuals, discusses potential limitations, and proposes solutions for practical applications. Section IV presents experimental results using the MSTAR Database \cite{b1} under various known and unknown target conditions, detailing the OOD detection score metrics and corresponding formulations. In Section V, we explore future research directions based on our approach. Finally, we conclude our findings in Section VI.

\begin{figure*}[htb!]
    \centering
    \begin{subfigure}[b]{0.1\textwidth}
        \centering
        \includegraphics[width=\textwidth]{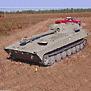}
    \end{subfigure}
    \begin{subfigure}[b]{0.1\textwidth}
        \centering
        \includegraphics[width=\textwidth]{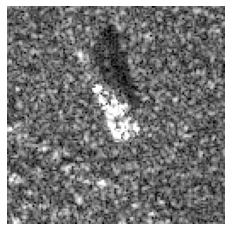}
    \end{subfigure}
    \begin{subfigure}[b]{0.1\textwidth}
        \centering
        \includegraphics[width=\textwidth]{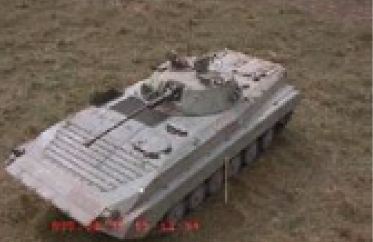}
    \end{subfigure}
    \begin{subfigure}[b]{0.1\textwidth}
        \centering
        \includegraphics[width=\textwidth]{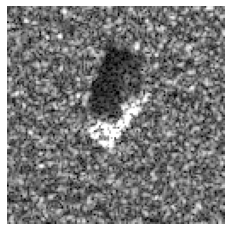}
    \end{subfigure}
    \begin{subfigure}[b]{0.1\textwidth}
        \centering
        \includegraphics[width=\textwidth]{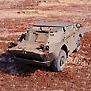}
    \end{subfigure}
    \begin{subfigure}[b]{0.1\textwidth}
        \centering
        \includegraphics[width=\textwidth]{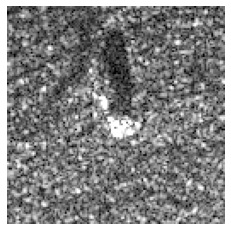}
    \end{subfigure}
    \begin{subfigure}[b]{0.1\textwidth}
        \centering
        \includegraphics[width=\textwidth]{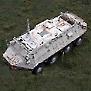}
    \end{subfigure}
    \begin{subfigure}[b]{0.1\textwidth}
        \centering
        \includegraphics[width=\textwidth]{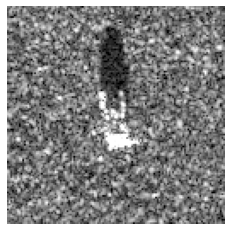}
    \end{subfigure}
    
    \vskip\baselineskip 
    
    \begin{subfigure}[b]{0.1\textwidth}
        \centering
        \includegraphics[width=\textwidth]{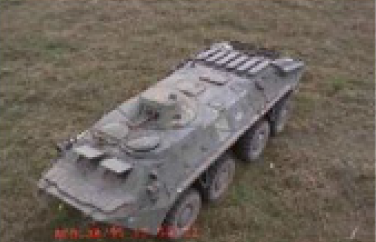}
    \end{subfigure}
    \begin{subfigure}[b]{0.1\textwidth}
        \centering
        \includegraphics[width=\textwidth]{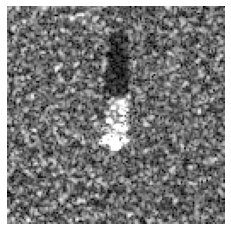}
    \end{subfigure}
    \begin{subfigure}[b]{0.1\textwidth}
        \centering
        \includegraphics[width=\textwidth]{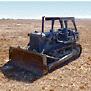}
    \end{subfigure}
    \begin{subfigure}[b]{0.1\textwidth}
        \centering
        \includegraphics[width=\textwidth]{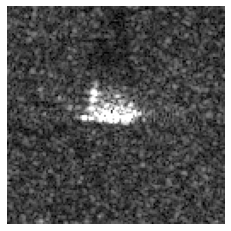}
    \end{subfigure}
    \begin{subfigure}[b]{0.1\textwidth}
        \centering
        \includegraphics[width=\textwidth]{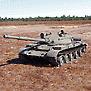}
    \end{subfigure}
    \begin{subfigure}[b]{0.1\textwidth}
        \centering
        \includegraphics[width=\textwidth]{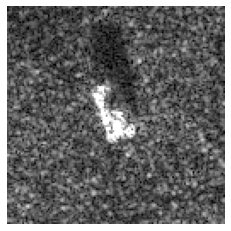}
    \end{subfigure}
    \begin{subfigure}[b]{0.1\textwidth}
        \centering
        \includegraphics[width=\textwidth]{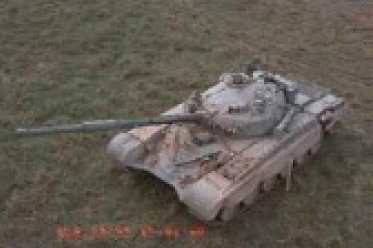}
    \end{subfigure}
    \begin{subfigure}[b]{0.1\textwidth}
        \centering
        \includegraphics[width=\textwidth]{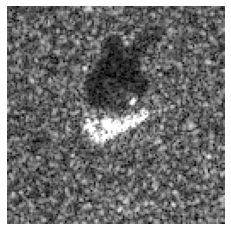}
    \end{subfigure}
    
    \vskip\baselineskip 
    
    \begin{subfigure}[b]{0.1\textwidth}
        \centering
        \includegraphics[width=\textwidth]{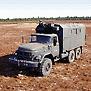}
    \end{subfigure}
    \begin{subfigure}[b]{0.1\textwidth}
        \centering
        \includegraphics[width=\textwidth]{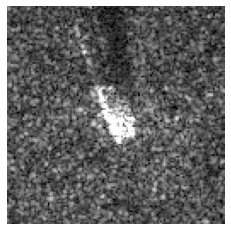}
    \end{subfigure}
    \begin{subfigure}[b]{0.1\textwidth}
        \centering
        \includegraphics[width=\textwidth]{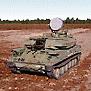}
    \end{subfigure}
    \begin{subfigure}[b]{0.1\textwidth}
        \centering
        \includegraphics[width=\textwidth]{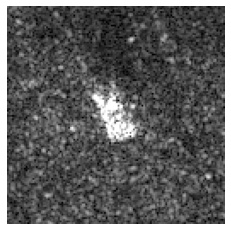}
    \end{subfigure}

    \caption{MSTAR Database: Optical images of military targets versus SAR images \cite{b1}. \\ Optical images of MSTAR database targets and their corresponding SAR images. \textit{Top to bottom rows, From left to right}: the targets are 2S1, BMP2, BRDM2, BTR60, BTR70, D7, T62, T72, ZIL, and ZSU.}
    \label{fig:MSTAR_images}
\end{figure*}

All experiments were conducted on hardware comprising an Intel Core i7-14700KF CPU with 64GB of memory and an NVIDIA GeForce RTX 3060 GPU with 12GB of memory. The software environment included a deep learning workstation running Microsoft Windows 11 Education and PyTorch 2.0.1 \cite{b56}.

\section{Motivation}

This section introduces the unique differences between SAR and optical images, along with the specific challenges these differences pose for OOD detection in SAR imagery. We also examine the informational limitations of post-processing-based OOD detection methods under diverse unknown target conditions. Following this, we propose a potential solution to enhance OOD detection in SAR imagery and demonstrate the suitability of our approach in overcoming these limitations. 

\subsection{Domain Gaps between Optical and SAR Images}

Outlier detection research has typically focused on optical images, which may not translate directly to SAR images. Figure 1 shows the comparison between the optical image and SAR image for each corresponding target class. Unlike optical images produced by reflected visible light, SAR images are generated through back-scattered radar signals \cite{b43, b44, b45}. This leads to less intuitive and more ambiguous patterns and textures, making it difficult to differentiate between classes without advanced processing techniques. Furthermore, SAR images are particularly prone to speckle and thermal noise, which exacerbate clutter issues and contribute to the high similarity between different target classes or regions \cite{b46, b47, b48}. As a result, modern computer vision methods often struggle to achieve the same level of performance in SAR image analysis as they do with greyscale image datasets and optical datasets, such as MNIST and CIFAR \cite{b50, b51}.

\subsection{Limitation of Post-Processing OOD Detection Methods}
The need for developing threshold optimization and unknown target detectors that are insensitive to the number and type of unknowns is illustrated in Figures 2 and 3. "AConvNet" \cite{b54} deep neural networks were trained with five known target classes from the MSTAR database, with the remaining class labels treated as unknown targets. Labels 1 through 10 correspond to the MSTAR Database targets '2s1,' 'bmp2,' 'brdm2,' 'btr60,' 'btr70,' 'd7,' 't62,' 't72,' 'zil131,' and 'zsu23,' respectively, in increasing order \cite{b1}. All label classes were included in the test.
Figure 2 illustrates the distribution of maximum softmax values (which serve as intuitive confidence scores) for 3,502 test samples, where labels 1 to 5 from the MSTAR database were used for training, while labels 6 to 10 were left untrained. The Y-axis represents the range of maximum softmax values, and the X-axis corresponds to the index of test target data, starting with targets from trained label classes. Blue dots indicate the distribution of maximum softmax values for test targets with trained labels, while orange dots represent the distribution for untrained label test targets. The horizontal dotted line represents the optimal threshold, which was estimated to achieve the highest accuracy for binary classification between known and unknown targets, using the maximum softmax values from all test data. As shown in Figure 2, the boundary between trained and untrained classes is well-defined, suggesting that the maximum softmax value serves relatively effectively as an outlier detector \cite{b22}. However, the results deteriorate significantly when the roles of trained and untrained label classes are reversed, as demonstrated in Figure 3. The boundary separating known from unknown targets becomes significantly less distinct, resulting in inefficient classification despite minimal change in the optimal softmax threshold, which suggests that softmax values are unreliable metrics for OOD detection. Given the variability of the softmax distribution under various unknown target conditions, feature-based OOD detection methods are expected to be highly sensitive to the characteristics of unknown targets. This issue is further exacerbated by the number of known and unknown targets, the domain gap between the training and test datasets, and the limitations of the training data. In other words, high performance in OOD detection for a particular setting does not guarantee similar performance across diverse unknown target conditions.

\begin{figure}[H]
    \centering
    \includegraphics[width=0.9\columnwidth]{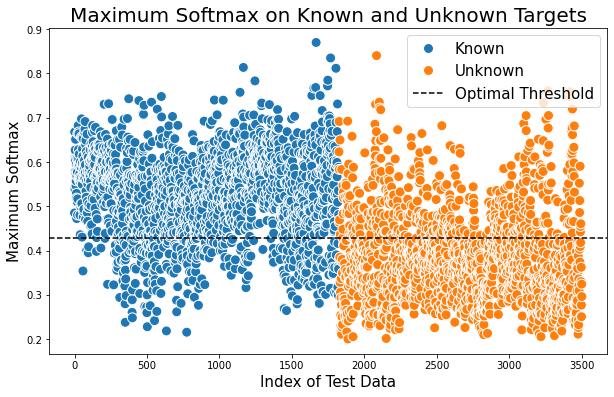}
    \captionsetup{justification=centering}
    \caption{Scatter Plot of MSP with Untrained Labels 6$\sim$10}
    \label{fig:softmax_cut}
\end{figure}

\begin{figure}[H]
    \centering
    \includegraphics[width=0.9\columnwidth]{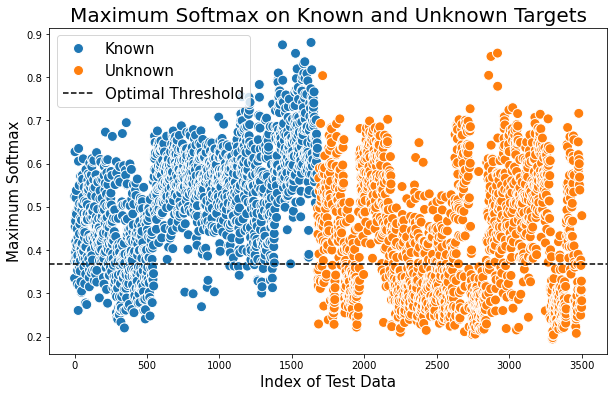}
    \captionsetup{justification=centering}
    \caption{Scatter Plot of MSP with Untrained Labels 1$\sim$5}
    \label{fig:softmax_cut}
\end{figure}

\subsection{Inside Out: Reversing From Central OOD to Peripheral OOD Clusters}

A potential solution to improve OOD detection robustness under varying target conditions is the use of classifier neural networks with adversarial learning \cite{b42, b65, b66}, applying domain generalization techniques. However, if training-phase-based OOD detection is not feasible and only post-processing OOD detection is available, reinterpreting feature information through our residual-based approach can offer valuable insights. 

In this regard, the transformation of two penultimate layer feature vectors into a standardized residual vector was proposed in \cite{b53}. This approach considers all possible combinations between two penultimate layer feature vectors, generating both in-class and inter-class residuals. This paper extends the work in \cite{b53} by further developing the mathematical formalism and providing guidance on how these properties can be applied for prior feature-based OOD detection, while building on the statistical reference space introduced in \cite{b53}.

The class-specific residual approach highlights differences between combinational pairs of two feature vectors (FVs), making in-class and inter-class residuals more explicit and amplifying variation than raw FVs. This enhances anomaly detection by increasing the relative distinction in raw FVs in feature scale. In addition, residuals between pairs can mitigate various types of noise, apart from the target signal, by subtracting two FVs that share common noise terms. While random noise, such as thermal noise, would not be canceled through subtraction, this method is particularly beneficial in SAR image classification, where natural clutters near the target area can degrade anomaly detection rates. In general, the clutter feature distribution is more distinguishable from the target feature distribution than individual feature information when considered as separate data points in latent space. If the target images are taken in similar scenes, the clutter signal distribution will follow a similar pattern if they originate from the same source of clutter. Generating residuals through the subtraction process cancels out the common clutter signals in both FVs. Hence, our approach offers robustness against variations in feature scales.

\begin{figure*}[htb!]
    \centering
    \begin{subfigure}[b]{0.45\textwidth}
        \includegraphics[width=\textwidth]{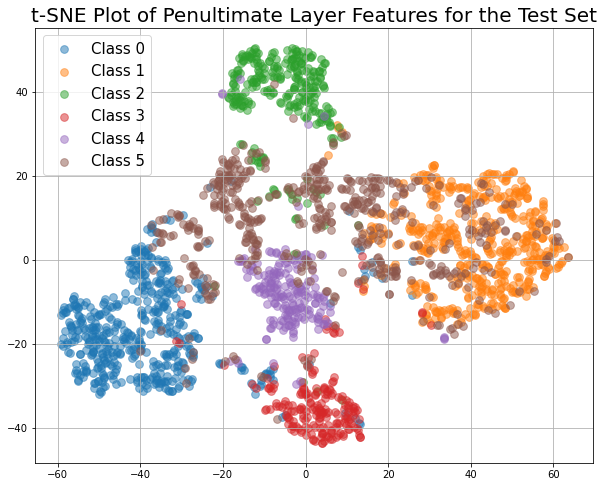}
        \caption{t-SNE Plot of Penultimate Layer Features: Central OOD Distribution with Surrounding ID Classes}
        \label{fig:tSNE_SPFV}
    \end{subfigure}
    \hfill
    \begin{subfigure}[b]{0.45\textwidth}
        \includegraphics[width=\textwidth]{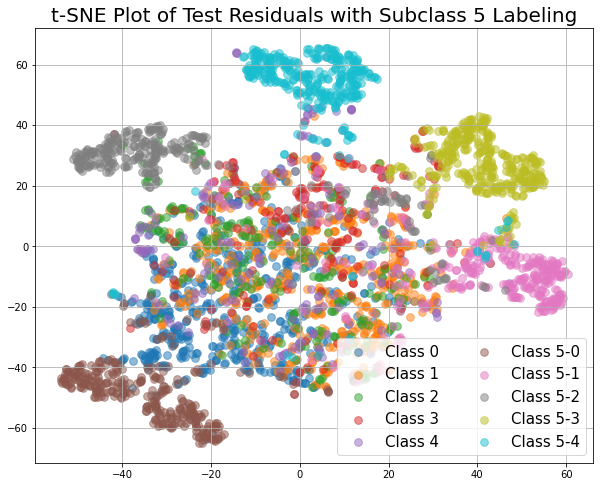}
        \caption{t-SNE Plot of Test Residuals: Clustered Peripheral OOD Detection with Centralized ID Distribution}
        \label{fig:tSNE_Residuals}
    \end{subfigure}
    \caption{Inverting ID and OOD Detection: From Central to Peripheral Detection Strategies}
    \label{fig: tSNE_comparison}
\end{figure*}

Direct usage of feature vectors can be sensitive to variations in scale and noise, potentially affecting anomaly detection performance. Previous post-processing methods using FVs are prone to the presence of natural clutter, speckle, and thermal noise. Their feature vectors based processing algorithms can be contaminated by these unwanted signals. On the other hand, generating residuals from FVs results in the loss of intrinsic feature information at specific vector locations. However, considering the purpose of an unknown target detector is to identify anomalies in the distribution as a binary classification rather than classify detailed semantic features with multiple labels, this residual interpretation approach might have an edge over direct feature interpretation. Interpreting the intrinsic meaning of each element in the FVs problem now turns into the problem of interpreting the probability distributional set. It actually converts from a feature-based OOD task into a re-interpreted field of OOD problems in probability density function (PDF) space. 

Figure 4 shows t-SNE visualizations of penultimate layer features (left) and class-specific residuals (right). In the left t-SNE plot, the colors blue, orange, green, red, and purple represent the trained in-distribution (ID) classes (labeled as classes 0, 1, 2, 3, and 4, respectively), while the brown points (labeled as class 5) denote untrained out-of-distribution (OOD) samples. Although class 5 may contain multiple unknown classes, they are grouped under a single label due to the uncertainty regarding the exact number of unknown classes. ID classes are arranged around the periphery, while the brown OOD points are dispersed throughout the center, partially overlapping with features from various ID classes. In this reference space, the OOD detection task involves distinguishing the central brown points from the surrounding ID points.

In contrast, the right t-SNE plot, based on class-specific residuals, displays a centralized cluster representing the in-class residuals of one ID class (independent of label), with five smaller clusters of inter-class residuals positioned around its periphery. Classes 0 through 4 denote the in-class residuals for each respective class, while Class 5-$i$ represents the residuals calculated between the test sample's feature vector and the feature vector of the known class labeled $i$. Unlike the previous approach, which addresses OOD detection using direct feature vectors, this method reframes the task by distinguishing the five peripheral OOD clusters from the central ID cluster. Although this approach requires analyzing each of the five OOD clusters individually, the localized clustering results in a clearer, more well-defined decision boundary for OOD detection.

In summary, while standardized residuals are computationally intensive, their benefits in mitigating noise, robustness against clutter, and intensive training with augmented data make them a superior choice for anomaly detection. By carefully capturing the underlying distributional patterns of standardized residuals for each class label, we can significantly enhance the effectiveness of unknown target detection.

\section{Method}
Figure 5 presents an overall flowchart for an unknown target detector, which will be explained throughout this section. Once the training data is collected, the SAR images and labels are prepared and input into the neural network classifier model for training. An all-convolutional networks (A-ConvNets) \cite{b54} is used for deep learning classifier models. It consists entirely of convolutional layers and is purported to be suitable for SAR images. In addition, we applied label smoothing in the loss function to prevent over-confidence and achieve a richer representation of feature vectors in the latent space \cite{b56}.

\begin{figure*}[htb!]
    \centering
    \includegraphics[width=0.8\textwidth]{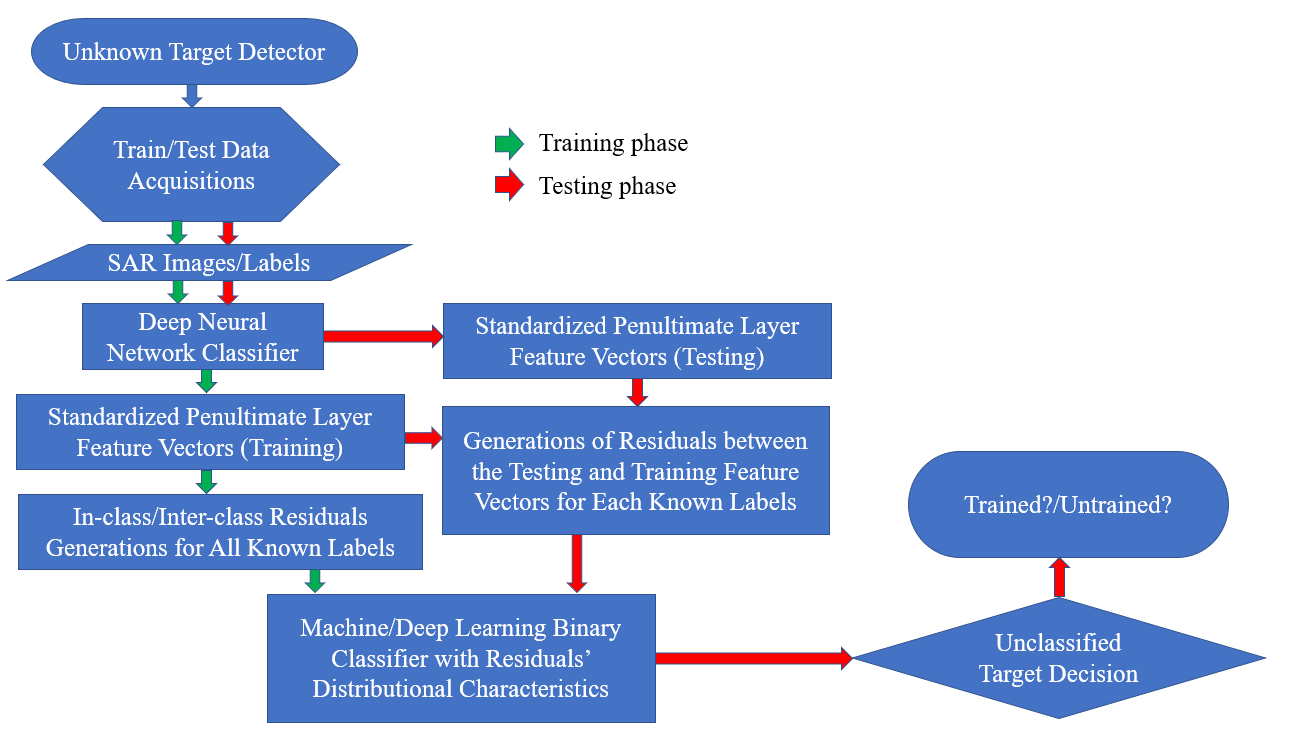}
    \captionsetup{justification=centering}
    \caption{Flow Chart of the Process}
    \label{fig:flow_chart}
\end{figure*}

We initially assessed the performance of our unknown target detector using the full MSTAR dataset. However, because most realistic scenarios involve limited training data, we also conducted experiments on a reduced version of the MSTAR dataset to simulate these conditions. The number of channels in the convolutional layers of the neural networks was adjusted to better accommodate our feature extraction requirements. Throughout this study, the neural network model was trained using an initial learning rate of \(5 \times 10^{-4}\), a learning rate scheduler with a step size of 100, a gamma decay rate of 0.1, for 50 epochs, and with a batch size of 32. The Adam optimizer was used throughout \cite{b55}.
Figures 5 and 6 depict the network architectures of 'AConvNet', with the SAR images having a width and height of 128.

Before the neural network computes the final logit/softmax values, feature vectors of length $N_{pen}$, where $N_{pen}$ represents the number of channels in the penultimate layer, are extracted. This penultimate layer captures high-level abstract features, making it useful for anomaly detection. $N_{pen}$, determined by the network architecture, represents the number of elements in each residual distribution, which will be explained later. It should be large enough for statistical analysis but balanced against computational cost. This penultimate layer contains concise and abstract feature information, often used for anomaly detection due to its richer representation of features than the logits/softmax layer \cite{b24, b26}.

Deep neural networks for SAR image classification, similar to those used for RGB images, are notorious for quickly overfitting the training data and becoming overconfident in test results based on training data features. The performance of discriminating the unclassified targets mainly relies on enriched feature representations during the training phase. As a result, the penultimate layers contain less variability, leading to poorer performance in detecting unknown targets. To give more variability to the softmax/penultimate layer feature vectors, we applied label smoothing, which is a technique that prevents the model from becoming overly confident in its predictions by assigning a small amount of probability mass to all classes, concentrating it solely on the true class. \cite{b60}

The smoothed label for class $i$, $\tilde{y}_i$, is given by

\begin{equation}
\tilde{y}_i = 
\begin{cases} 
1 - \alpha + \frac{\alpha}{K} & \text{if } i = c \\
\frac{\alpha}{K} & \text{if } i \neq c 
\end{cases}
\end{equation}
where $\alpha$ is the smoothing parameter that controls the degree of smoothing for each class label, ranging from 0 with no smoothing to 1 with uniform smoothing, where $\alpha = 0.4$ is adopted in this paper. $K$ is the total number of class labels, $i$ is the index for a certain class, and $c$ is the ground truth class index.

By applying label smoothing to the loss function, the training process will yield less over-confidence and, hence, more learnable distributions for the penultimate layers.

\subsection{Class-specific Residual between Trained and Test Samples}
After the model training is finished, the penultimate layer feature vectors are obtained for all correctly classified training data (hence, for all known labels). After standardizing these feature vectors, we construct the reference basis for known/unknown target spaces by generating their residuals. Standardization ensures that residuals can be meaningfully compared across different classes.

Our aim in this study is to develop a more reliable and stable unknown target detector, irrespective of the number/type of known/unknown samples. This objective directly corresponds to the statistical analysis task, necessitating standardized metrics. In a PyTorch implementation \cite{b56} with Batch Normalization \cite{b57}, the standardization of the penultimate layer's outputs is approximate, as Batch Normalization includes learnable scaling and shifting parameters. Manual standardization needs to be performed to ensure precise standardization of the penultimate layer's output vectors. The Standardized Penultimate Feature Vector (SPFV) formulation for two different indices $i$ and $j$ in the components of the penultimate layer is given by

\begin{equation}
\mathbf{v}_i = \frac{\mathbf{x}_i - \mu_i}{\sigma_i}, \quad \mathbf{v}_j = \frac{\mathbf{x}_j - \mu_j}{\sigma_j}
\end{equation}

Our idea of using residuals stems from the fundamental nature of statistics. When the observed data samples are collected from their true model $f(x_{i},\Theta )$, their normalized residuals are expected to be Gaussian random noise. This allows us to frame OOD detection as a statistical residual analysis problem. Statistical theory in this frame is given by

\begin{equation}
\chi ^{2}_{i}=\frac{y_{i}-f(x_{i},\Theta )}{\sigma _{i}}
\end{equation}

\begin{equation}
\chi ^{2}\sim \mathit{N}(0,1)
\end{equation}

where $\chi$ is a set of chi-square test statistics for the entire sampled data, $\chi^2_{i}$ is the $i$-th standardized test statistic, $y_{i}$ is the $i$-th experimental measurement, $f(x_{i}, \Theta)$ is the corresponding theoretical value (or true value), $\Theta$ represents the parameters of the theoretical function, and $\sigma_{i}$ is the $i$-th standard deviation of the experimental measurement. Equation (1) cannot be directly applied to the target classification task to determine if the test samples match any of the classes in the training samples because there is no theoretical target characteristic function, and the uncertainty for each test characteristic is not provided.

These problems stem from: (1) training data samples not being true models of the given class because they do not fully represent all configurations of features with a limited number of trained datasets, (2) a domain gap between training data samples and test data samples due to measurement conditions and systematic errors during measurement, and (3) the function that describes image classification having an unknown degree of freedom.

An alternative remedy for mimicking the normalized metrics is calculating the residuals between classes and observing their distributions.

Instead, our approach uses the residual vectors between the same class and different classes of SPFVs. \cite{b61} The residual metrics can be calculated using all possible combinations of each feature between all training samples.

\begin{equation}
\begin{aligned}
\mathbf{r}_{ij} &= \left \{ \mathbf{v}^{m}_{i} - \mathbf{v}^{n}_{j} \right \}_{\left \{ i = {1, ..., N_{m}}, j = {1, ..., N_{n}} \right \}}^{\left \{ m, n = {1, ..., C} \right \}} \\
&\text{with} \quad (i \neq j \text{ when } m = n)
\end{aligned}
\end{equation} 

\begin{figure*}[htb!]
    \centering
    \includegraphics[width=0.8\textwidth]{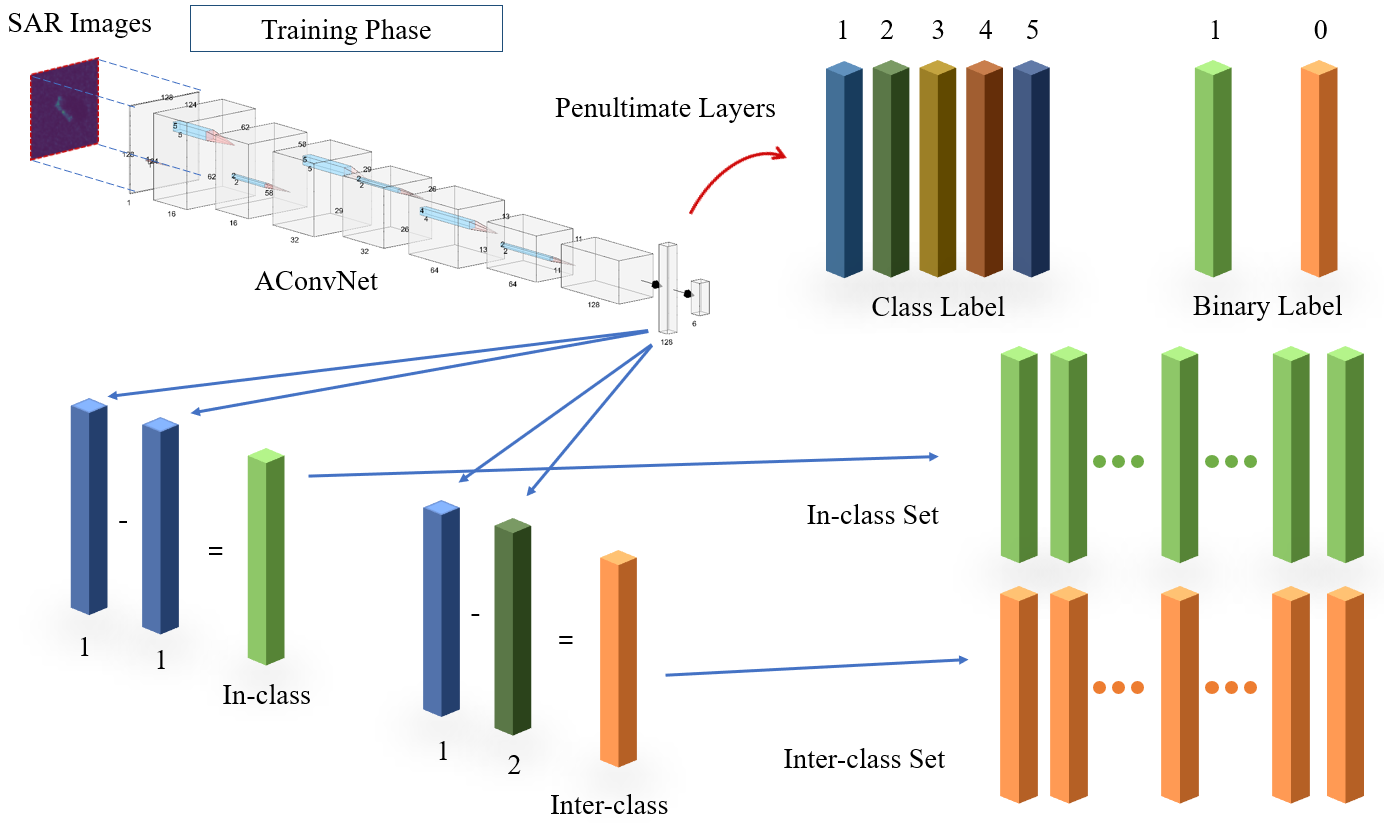}
    \captionsetup{justification=centering}
    \caption{Training Phase: Generations of In-class and Inter-class Residuals}
    \label{fig:flow_chart}
\end{figure*}

\begin{figure*}[htb!]
    \centering
    \includegraphics[width=0.8\textwidth]{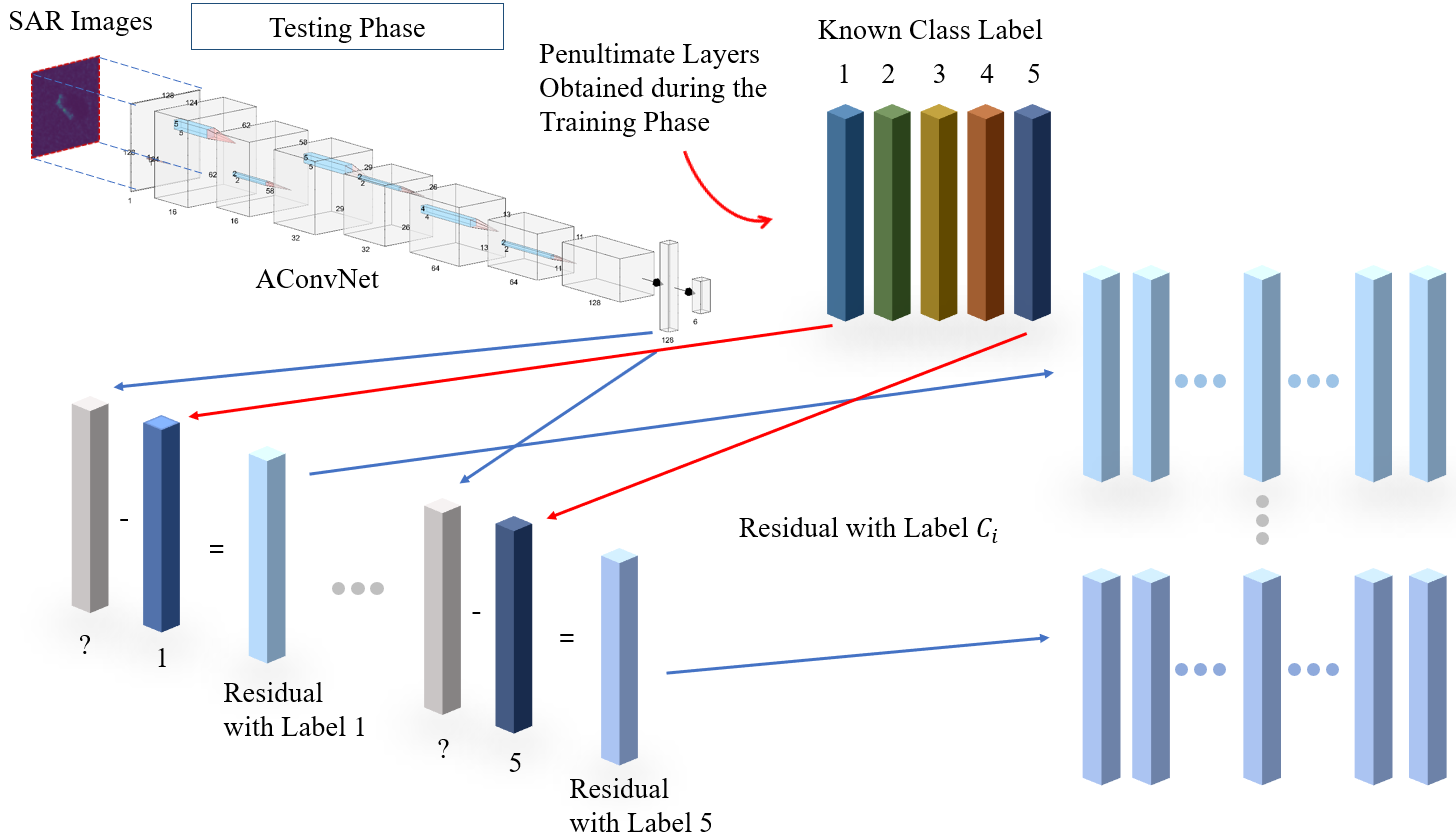}
    \captionsetup{justification=centering}
    \caption{Testing Phase: Generation of Residuals with label $C_{i}$}
    \label{fig:flow_chart}
\end{figure*}

where $C$ is the total number of known label classes, $m$ and $n$ are the indices of the class labels that run from 1 to $N_{c}$, $N_{m}$ and $N_{n}$ are the total numbers of data for the $m$-th and $n$-th indexed class labels, respectively, $i$ and $j$ are the indices representing specific data of the $m$ and $n$ class labels, respectively, and $v^{m}_{i}$ and $v^{n}_{j}$ are the first and second SPFVs in the residual formulation. We call the residuals between the same class labels in-class residuals (\(m = n\)) and those between different class labels inter-class residuals (\(m \neq n\)). This process is demonstrated in Figure 6.
In this formalism, each residual calculation provides one distributional data point regardless of the location of features in the latent space. In other words, the residual is calculated along the feature vector (element-wise subtraction), not across the feature. In another formalism, the residual can be calculated across the feature, where the length of each residual vector is the number of data points. However, this formalism will not be discussed in this paper.

With this setup, the total number of in-class residuals (where \(m = n\)) is given by
\begin{equation}
\sum_{m=1}^{C} \binom{N_m}{2} = \sum_{i=1}^{C} \frac{N_m (N_m - 1)}{2}
\end{equation}

The total number of inter-class residuals (where \(m \neq n\)) is given by
\begin{equation}
\sum_{1 \leq m < n \leq C} N_m \times N_n
\end{equation}

The total number of pairs, when the total number of train data is $N_{t}$, regardless of class, is given by
\begin{equation}
\binom{N_{t}}{2} = \frac{N_{t}(N_{t}-1)}{2} \quad \text{where} \quad N_{t} = \sum_{m=1}^{C} N_m
\end{equation}

During the test phase, when a single test SAR image is entered into the pre-trained deep neural network classifier, the corresponding SPFV is obtained. This SPFV combines with all training SPFVs across all classes to generate residuals. When the test SPFV combines with the \(m\)th index of a class label to generate \(N_{m}\) residuals, we refer to these as test residuals with label \(m\). This step goes through \(m = 1\) to \(m = C\), which is the total number of all known classes from the training phase. A detailed test residual generation is shown in Figure 7.

\begin{equation}
\mathbf{r}^{\text{test}}_i = \left \{ \mathbf{v}^{\text{test}} - \mathbf{v}^{\text{train}, m}_i \right \}_{\left \{ i=1, ..., N_m \right \}}^{\left \{ m=1, ..., C \right \}}
\end{equation}

Then, the total number of test residuals is simply the multiplication of the number of the total test data \textit{M} by the number of the total number of training data, which is given by
\begin{equation}
\text{Total test residuals} = M \times \sum_{m=1}^{C} N_m
\end{equation}

In-class, inter-class, and the total number of residuals grow quadratically from the original number of $N_{t}$ penultimate layer feature vectors. From a data augmentation perspective, this method might provide a richer representation of training data variation than feature vectors' direct usage. Whether these increased numbers of residuals provide unique and useful information for OOD detection and how to construct the residual pattern recognition should be carefully discussed. Quadratically augmented training data might be more suitable for training with deep neural networks than machine learning models.

To mitigate the computational time during the test phase for practical application, $N_{t}$ could be substituted with a certain sample number or sampling rate proportion to the original $N_{t}$ while not significantly affecting the accuracy of unknown target detection rate.

Using standardized residuals from SPFVs is relatively more computationally intensive than directly applying SPFVs, as it considers all possible pairwise combinations. However, this approach generates more diverse and augmented training data, reducing the risk of overfitting in limited data scenarios, despite some expected redundancy of information in the generated residuals. The primary advantage of the residual approach lies in its ability to utilize quadratically augmented residual data for robust training. In practical terms, computation time refers to the decision-making duration during the test phase, where extended training time remains acceptable. Sampling in-class and inter-class residuals during the testing phase presents a potential solution for reducing computational load. While reduced sampling rates may impact OOD detection performance, especially in scenarios with limited training data, the advantages of residual transformation can still be retained if the underlying OOD detection algorithm effectively analyzes both in-class and inter-class residual patterns. Within this algorithm pipeline, this approach achieves computational efficiency comparable to direct SPFV applications without compromising accuracy. However, caution is necessary, as reduced sampling rates may limit the practical benefits of the residual-based method when training data is constrained.

\subsection{Construction of Reference Space with Residuals' Characteristics}

As discussed in the previous section, the residuals calculated from SPFV do not necessarily follow a normal distribution with a mean of 0 and variance of 1.
\begin{equation}
\mathbf{r}_{ij} \not\sim \mathit{N}(0,1)
\end{equation}
In addition, the extent to which the residuals follow the Gaussian distribution is not an effective metric for distinguishing between in-class and inter-class residuals. Due to the shared features between classes, inter-class residuals can also show a similar Gaussian distribution shape to that of in-class residuals. Since each SPFV is standardized, its mean values are zero, and its variances are 1.

\begin{equation}
\mu_{\mathbf{v_i}} = \mathbb{E}[\mathbf{v}_{i}] = \mu_{\mathbf{v_j}} = \mathbb{E}[\mathbf{v}_{j}] = 0
\end{equation}

\begin{equation}
\sigma_{\mathbf{v_i}}^2 = \mathbb{E}[(\mathbf{v}_{i} - \mathbb{E}[\mathbf{v}_{i}])^{2}] \\
= \mathbb{E}[\mathbf{v}_{i}^{2}] = 1
\end{equation}

\begin{equation}
\sigma_{\mathbf{v_j}}^2 = \mathbb{E}[(\mathbf{v}_{j} - \mathbb{E}[\mathbf{v}_{j}])^{2}] \\
= \mathbb{E}[\mathbf{v}_{j}^{2}] = 1
\end{equation}

Since the residual is calculated by taking a difference between two SPFVs, we have
\begin{equation}
\mathbf{r}_{ij} = \mathbf{v}_i - \mathbf{v}_j
\end{equation}

Then, the mean value of the residual is the expectation value of the residual, which is given by

\begin{equation}
\begin{aligned}
\mu_{\mathbf{r}_{ij}} &= \mathbb{E}[\mathbf{r}_{ij}] = \mathbb{E}[\mathbf{v}_i - \mathbf{v}_j] \\
&= \mathbb{E}[\mathbf{v}_i] - \mathbb{E}[\mathbf{v}_j] = \mu_{\mathbf{v}_i} - \mu_{\mathbf{v}_j} = 0
\end{aligned}
\end{equation}

The variance of the residuals is the expectation value of the square of the difference between two SPFVs, which can be expressed as follows

\begin{equation}
\begin{aligned}
\sigma_{\mathbf{r}_{ij}}^2 &= \mathbb{E}[(\mathbf{r}_{ij}-\mathbb{E}[\mathbf{r}_{ij}])^{2}] = \mathbb{E}[\mathbf{r}_{ij}^{2}] \\
&= \mathbb{E}[(\mathbf{v}_i - \mathbf{v}_j)^2] \\
&= \mathbb{E}[\mathbf{v}_i^2] + \mathbb{E}[\mathbf{v}_j^2] - 2\mathbb{E}[\mathbf{v}_i \mathbf{v}_j] \\
&= 2(1 - \rho_{ij})
\end{aligned}
\end{equation}

where the Pearson's correlation coefficient $\rho$ is given by 

\begin{equation}
\rho_{ij} = \mathbb{E}[\mathbf{v}_i \mathbf{v}_j] = \frac{\text{Cov}(\mathbf{v}_i, \mathbf{v}_j)}{\sigma_{\mathbf{v}_i} \sigma_{\mathbf{v}_j}} = \text{Cov}(\mathbf{v}_i, \mathbf{v}_j)
\end{equation}

Pearson's correlation coefficient $\rho$ ranges from 1 to -1 and cannot exceed this range due to Cauchy-Swarch inequality (from 1 to perfect positive correlation, 0 no correlation, and -1 perfect negative correlation). This translates to the variance of the residual ranging from 0 to 4.

For in-class residuals, Pearson's correlation coefficient is expected to be close to 1 since paired the same class feature vectors will be alike, while the coefficient is expected to be away from 1 due to their feature vectors being dissimilar. Although Pearson's correlation coefficient is calculated to obtain the variance value, this metric serves as another metric from variance.

Assuming that the measurement error for each feature component varies insignificantly allows us to scale it using a single numerical measurement error, as shown in equation (3), which in turn supports equation (4). This assumption further constrains the normal distribution-like characteristics of the in-class residual distribution, partially similar to \cite{b24}, but in the context of residuals. For example, if the feature vectors differ significantly between label classes, the inter-class residual will deviate more from equation (4), resulting in a flattened distribution.

Figure 8 shows the residuals between the trained labels and a test sample for each of the five trained classes. In the figure, the test sample class is a '2S1 gun,' which the model has been trained on (the MSTAR image of the '2S1 gun' is shown at the bottom right). In each of the five histograms, the blue color represents the \(m\)th label train SPFV histogram (where the \(m\)th label corresponds to ’2S1’, ’BMP2’, ’BRDM2’, ’BTR-60’, ’BTR-70'), the red color represents the '2S1 gun' labeled test SPFV histogram, and the green color represents their residuals. All of the green residual histograms have mean values close to 0 with varying variances. While the others exhibit flattened distributions with a variance of more than 1, only the first histogram shows a distribution with a variance of less than 1. If the class label of the test sample were not one of the trained label classes, none of the green residuals would exhibit such behavior, allowing us to identify the test sample as an unknown target. In-class residuals typically have smaller variance and a Pearson's correlation coefficient close to 1, while inter-class residuals have larger variance and a Pearson's correlation coefficient close to or below 0. In short, we use the statistical properties of residuals to distinguish between classes, particularly to determine whether test data belongs to a known class or is an anomaly/outlier (unknown class).

\begin{figure*}[htb!]
	\includegraphics[scale=0.4]{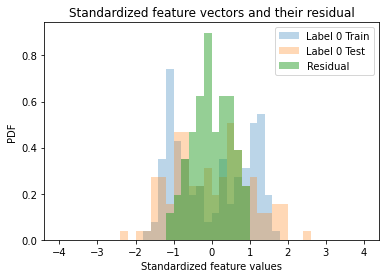}
	\includegraphics[scale=0.4]{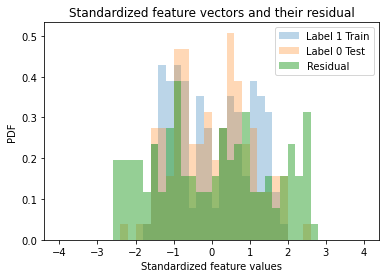}
    \includegraphics[scale=0.4]{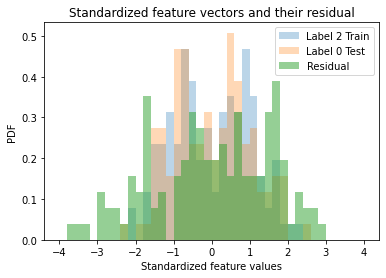}
	\includegraphics[scale=0.4]{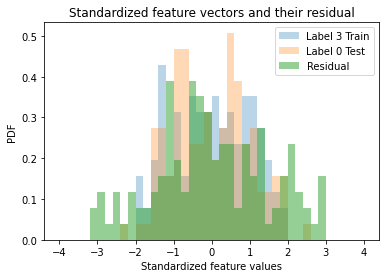}
	\includegraphics[scale=0.4]{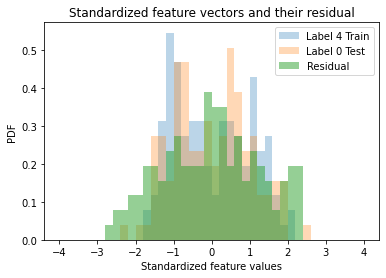}
	\includegraphics[scale=0.4]{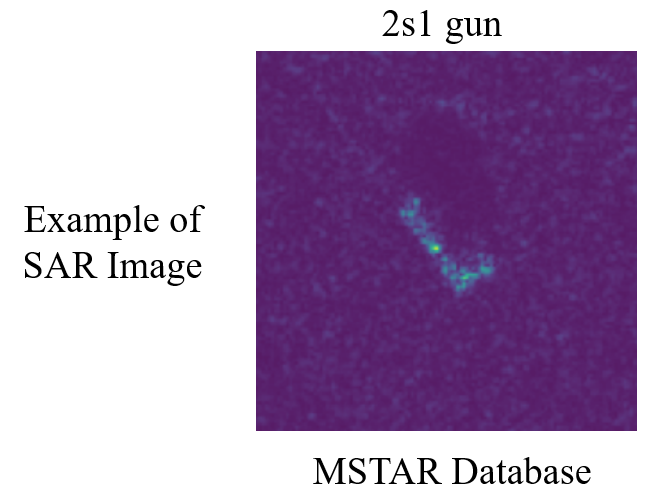}
    \caption{Residuals for the pre-trained target classes '2s1' in experiment 1 \textit{(first)} Residual for matched test class
    \textit{(the rest four)} Residual for mismatched test class}
    \label{fig: Histogram comparison}
\end{figure*}

\subsection{Out-of-Distribution Methods with Class-specific Residuals}
In order to compare the relative performance between feature-based and class-localized residual-based OOD detection approach, (1) statistical distance (2) OpenMax \cite{b26} (3) Mahalanobis distance \cite{b24} (4) ViM \cite{b31} (5) kNN (k-Nearest Neighbors) OOD detection methods are experimented as examples for original feature-based OOD detection methods. Each feature-based OOD detection methods serve as different aspects to evaluate the suitability of transformation of feature vectors into residual pairs in our approach.

\subsubsection{Statistical distance}

This method could be the simplest and most straightforward method to separate OOD from ID based on the statistical distributional characteristics of in-class residuals. This method calculates the in-class residuals and their corresponding skewness, kurtosis, variance, and Pearson's correlation coefficient (\(\rho\)) by labels.
Skewness and kurtosis are given by

\begin{equation}
\text{skewness} = \frac{1}{n} \sum_{i=1}^{n} \left( \frac{r_i - \mu_r}{\sigma_r} \right)^3
\end{equation}

\begin{equation}
\text{kurtosis} = \frac{1}{n} \sum_{i=1}^{n} \left( \frac{r_i - \mu_r}{\sigma_r} \right)^4
\end{equation}

The 'statistical distance' can be calculated using a combination of four statistical characteristics: variance, $1 -\rho$ (to align $\rho$ with other satistics), skewness, and kurtosis. While these metrics may lack robustness across diverse scenarios, we include them here as an exemplary set for comparison purposes. However, it remains uncertain which statistics might enhance or diminish accuracy in distinguishing known from unknown target classes, so we experimented with each individual statistic as well as their combinations. Our results indicate that higher-order statistics, such as skewness and kurtosis, are ineffective for direct unknown target detection. When using residuals, defining distance by variance alone, $1 -\rho$ alone, or a combination of variance and $1 -\rho$ yielded comparable accuracy across various unknown target conditions, making it challenging to identify a single optimal statistic. Consequently, we select the statistical distance $d_{\text{stat}}$ as the sum of variance and $1 -\rho$, with uniform weighting, as an illustrative example.

\begin{equation}
d_{\text{stat}} = 1 - \rho + \sigma
\end{equation}

In this set-up, the feature based OOD detection method is diffult to be applied as feature vector itself's variance is a poor metric for OOD detection, and pearson's coefficient can be calculated between two vectors, not from the single vector. Instead, feature correlation based statistical distance method is applied, which is similar to our class-specific residual approach. However, pearson's coefficients are directly calculated in all-combinational pairs for in-class and inter-class feature vectors unlike they are calculated for in-class and inter-class residual vectors. The comparison between two these methods can demonstrate the residual themselves still contain the correlational information between features after the transformation.
\subsubsection{OpenMax}
OpenMax fits a Weibull model to the distances between training feature vectors and each class’s activation center, modeling a boundary for each class in the OpenMax layer. During testing, the distance between a test sample's feature vector and each class activation center is calculated, and confidence scores are derived from the Weibull model to estimate the likelihood of the sample belonging to an unknown class. This modifies softmax outputs by replacing them with adjusted probabilities, including an unknown category. \cite{b26} By comparing the feature-based OpenMax with a modified algorithm that uses class-specific residuals as inputs, the improvement observed with the residual-based approach suggests that statistical model-based residual learning may hold promise for future research directions.

\subsubsection{Mahalanobis Distance}
Mahalanobis distance based OOD detection method shares partial similarities with our approach. Mahalanobis distance assumes that the feature distribution of each class is Gaussian. It calculates the Mahalanobis distance between a test sample's feature representation and the mean feature vector of each training class in multivariate space, factoring in the covariance of the training data. This covariance-adjusted distance is used as an OOD detection metric, with larger distances indicating higher likelihoods of the sample being OOD \cite{b24}. This feature-based OOD detection method is particularly shows outstanding performance for feature based OOD detection across various OOD datasets \cite{b31, b66, b67}. The improvement with transformation into the residual bases as inputs could imply our residual base is more robust as reference space for OOD/ID boundary.

\subsubsection{ViM}
ViM applies PCA to the training feature vectors to define a principal subspace representing the in-distribution data. For a test input, it calculates the residual by measuring the norm of the difference between the test feature vector and its projection onto this principal subspace. This residual norm is scaled by a factor, alpha, computed as the ratio of the average maximum logit to the average training residual norm. The scaled residuals are then appended to the test logits as virtual logits. Finally, applying softmax to these augmented logits yields a ViM score, which serves as an OOD metric. This OOD detection method uses multiple processing steps on feature vectors, where the original intrinsic information of the features may play a key role, making residuals as inputs potentially unsuitable for application with ViM. Furthermore, decomposing training inputs and processing test inputs with the pre-processed training data, as in the case of distributional data types like our class-localized residuals, may also be unsuitable.

\subsubsection{kNN Distance/Density}
The kNN algorithm is particularly useful for estimating similarities and density in feature space, providing an intuitive approach to assess whether a test sample belongs to the in-distribution data. In this approach, the density of clusters within each class can be measured by evaluating the similarity of test inputs to mean representations of in-distribution training data within a class-specific residual space. While kNN algorithm allows for flexible distance metrics, our experiment uses the simple Euclidean distance to compare OOD metrics between two types of inputs. By using this straightforward metric, we aim to establish whether a basic approach can enhance OOD detection in our residual-based method. If Euclidean distance proves insufficient for improving detection rates, this would indicate the need for more refined processing techniques as a direction for future development.
 
\begin{table*}[htbp]
\centering
\caption{\textsc{TRAINING AND TEST SAR SAMPLES IN MSTAR DATABASE}}
\begin{tabular}{c c c c c c c |c c c c c|c}
\hline
\hline
& \multicolumn{11}{c}{MSTAR Database} \\ \cline{2-13}
Exp. & Train & 1 & 1 & 1 & 1 & 1 & 2 & 2 & 2 & 2 & 2 & Test \\ \hline
Case & Train & 1 & 2 & 3 & 4 & 5 & 1 & 2 & 3 & 4 & 5 & Test \\ \hline
Data size & 100\% & 100\% & 50\% & 25\% & 10\% & 5\% & 25\% & 25\% & 25\% & 25\% & 25\% & 100\% \\ \hline
Dep. Angle & 14-16 & - & - & - & - & - & - & - & - & - & - & 17 \\ \hline
2S1 & 299 & 299 & 150 & 75 & 30 & 15 & 75 & 0 & 75 & 0 & 75 & 573 \\ 
BMP2 & 698 & 698 & 349 & 175 & 70 & 35 & 175 & 0 & 0 & 0 & 175 & 587 \\ 
BRDM2 & 298 & 298 & 149 & 75 & 30 & 15 & 75 & 0 & 75 & 75 & 75 & 274 \\ 
BTR60 & 233 & 233 & 117 & 59 & 24 & 12 & 59 & 0 & 59 & 59 & 0 & 196 \\ 
BTR70 & 256 & 256 & 128 & 64 & 26 & 13 & 64 & 0 & 0 & 0 & 0 & 195 \\ 
D7 & 299 & 0 & 0 & 0 & 0 & 0 & 0 & 75 & 0 & 75 & 0 & 274 \\ 
T62 & 299 & 0 & 0 & 0 & 0 & 0 & 0 & 75 & 75 & 75 & 0 & 273 \\ 
T72 & 691 & 0 & 0 & 0 & 0 & 0 & 0 & 173 & 173 & 173 & 0 & 582 \\ 
ZIL131 & 299 & 0 & 0 & 0 & 0 & 0 & 0 & 75 & 0 & 75 & 0 & 274 \\ 
ZSU23 & 299 & 0 & 0 & 0 & 0 & 0 & 0 & 75 & 0 & 75 & 0 & 274 \\ \hline
Total & 3671 & 1784 & 893 & 448 & 180 & 90 & 448 & 473 & 457 & 607 & 325 & 3502 \\ \hline
\end{tabular}
\end{table*}

\section{Result}

\subsection{Datasets and Experimental Set-up}

For the experiments, we used the MSTAR database with the label names ['2s1', 'bmp2', 'brdm2', 'btr60', 'btr70', 'd7', 't62', 't72', 'zil131', 'zsu23'], which will be referred to as label numbers 0 to 9, respectively, in ascending order. In Experiment 1, the training data size varies from 5$\%$ to 100$\%$ while keeping the unknown target types (labels 5 to 9) constant. We performed into five different cases based on the percentage of training data: (1) 5$\%$, (2) 10$\%$, (3) 25$\%$, (4) 50$\%$, and (5) 100$\%$. Table 1 shows the MSTAR database with different percentages of training data and their corresponding total number of training targets. In Experiment 2, the number and types of unknown targets are varied while the dataset size is fixed at 25$\%$. The five cases are as follows: (1) labels 5, 6, 7, 8, and 9 are not trained, (2) labels 0, 1, 2, 3, and 4 are not trained, (3) labels 1, 4, 5, 8, and 9 are not trained, (4) labels 0, 1, and 4 are not trained, and (5) labels 3 to 9 are not trained. Experiment 2 compares the unknown target detection accuracy across varying known/unknown target classes for each model. The summarized set-up of experiments is shown in Table II.

\subsection{Performance Metrics}

In the computer vision domain, metrics such as the Area Under the Receiver Operating Characteristic Curve (AUROC) and the False Positive Rate at 95\% True Positive Rate (FPR95) are commonly preferred for their robustness and threshold independence \cite{b49, b22}. These metrics assess performance across decision thresholds, addressing known-unknown imbalances. However, due to the nature of our proposed feature residual scoring system, which employs multiple class-specific OOD scores and thresholds, metrics like AUROC and FPR95 may not provide a effective comparison to single-threshold OOD methods. Additionally, in real-world scenarios, users must develop algorithms to optimize threshold settings, which should be considered in performance evaluations for unknown target detection. In the experiments conducted in this paper, we compare our residual-based reshaped OOD methods with original feature-based methods. In addition to AUROC and FPR95, we report total accuracy, known target accuracy, and unknown target accuracy using a predefined threshold based on the training dataset, providing a straightforward and intuitive assessment of detector performance.

\subsubsection{Accuracies}

Overall Accuracy (OA) is defined as the total number of correct classifications divided by the total number of test data. Known Accuracy (KA) is defined as the correct classification of known targets divided by the total number of known test data. Unknown Accuracy (UA) is defined as the correct classification of unknown targets divided by the total number of unknown test data, which are given by

\begin{equation}
\text{OA} = \frac{\text{Total Correct Classifications}}{\text{Total Test Data}}
\end{equation}

\begin{equation}
\text{KA} = \frac{\text{Correct Known Target Classifications}}{\text{Total Known Test Data}}
\end{equation}

\begin{equation}
\text{UA} = \frac{\text{Correct Unknown Target Classifications}}{\text{Total Unknown Test Data}}
\end{equation}
\comment{
We evaluate our model's performance using two key metrics: \textbf{AUROC} and \textbf{FPR95}.

\textbf{AUROC (Area Under the Receiver Operating Characteristic Curve)}: AUROC represents the area under the ROC curve, which plots the true positive rate (TPR) as a function of the false positive rate (FPR) at different classification thresholds. Mathematically, AUROC can be expressed as the following integral over the ROC curve:

\begin{equation}
\text{AUROC} = \int_{0}^{1} \text{TPR}(\text{FPR}) \, d(\text{FPR})
\end{equation}

where \(\text{TPR}(\text{FPR})\) represents the true positive rate as a function of the false positive rate.

\textbf{FPR95 (False Positive Rate at 95\% True Positive Rate)}: FPR95 is defined as the false positive rate at the threshold where the true positive rate (TPR) is 95\%. This metric captures the proportion of out-of-distribution samples that are incorrectly classified as in-distribution when the model achieves a sensitivity level of 95\% for in-distribution data. Formally,

\begin{equation}
\text{FPR95} = \text{FPR at} \, \text{TPR} = 0.95
\end{equation}

Thus, FPR95 is obtained by identifying the threshold on the ROC curve that corresponds to a TPR of 0.95 and reporting the corresponding FPR at this threshold.
}

\subsubsection{Sampling}
Since the residual-based approach can generate all possible combinations of test feature vectors with training feature vectors from known class labels, we sample only 10\% of the training feature vectors to reduce computational costs. However, this sampling rate may present challenges in Experiment 1 with the reduced dataset, particularly when in-class residuals are used, as the effectiveness of residual approaches depends on capturing intricate patterns within a larger, augmented data set.
\subsubsection{Thresholds}
A user-defined percentile is specified for each class as a threshold to differentiate between known and unknown classes. Typically, threshold values are determined based on the distribution of the training dataset, such as the 95th or 100th percentile, under the assumption that values not encountered during training are likely to be OOD. However, this thresholding approach may be less effective when training data is limited. Once the backbone model is selected, we determine class-specific thresholds by setting them to the upper percentile (100\%) of the in-distribution data for each known class label.

\begin{figure*}[htb!]
    \centering
    \includegraphics[width=0.8\textwidth]{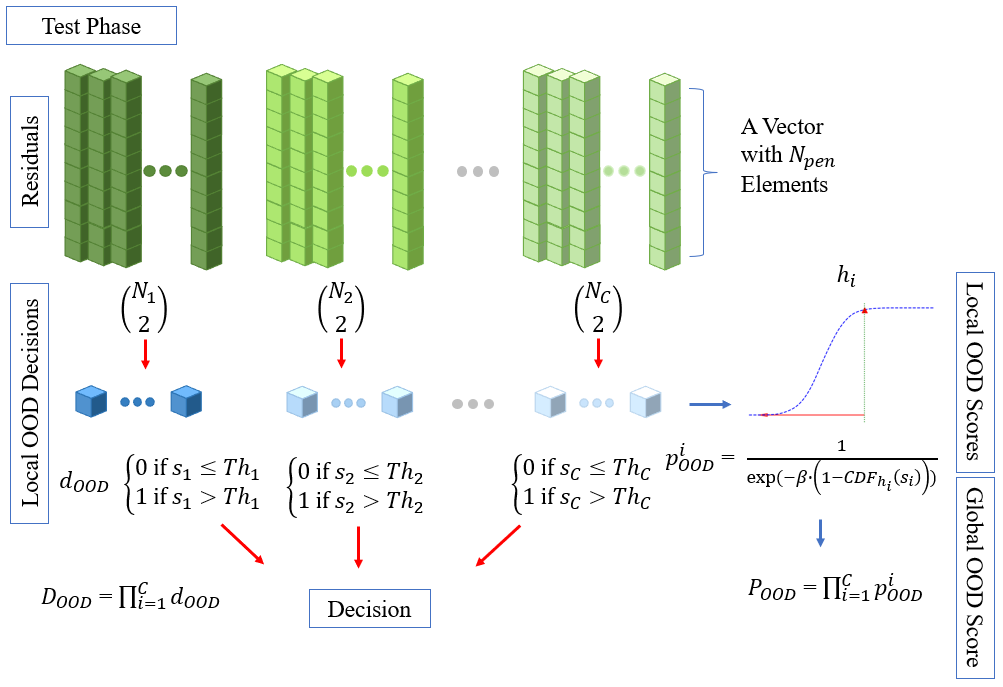}
    \captionsetup{justification=centering}
    \caption{Global OOD Score/Decision Estimation Steps}
    \label{fig:flow_chart}
\end{figure*}
\subsubsection{Global OOD Decision}

Figure 9 illustrates the schematic process for deriving the global OOD score and making a decision based on class-specific local OOD metrics. Due to the class-localized nature of residuals, each test input feature generates a residual in combination with each known class, resulting in $C$ local OOD metrics from the detection method. The distributions $h_{1}$, $h_{2}$, ..., $h_{C}$ represent class-specific in-distribution (ID) metrics, derived from the training dataset to serve as reference spaces for each known class. The test metrics, $s_{1}$, $s_{2}$, ..., $s_{C}$, are the OOD scores generated for the test input, while the thresholds $Th_{1}$, $Th_{2}$, ..., $Th_{C}$ are user-defined based on the respective $h_{1}$, $h_{2}$, ..., $h_{C}$ distributions. 
Each local OOD decision, $d_{OOD}$, is determined by whether the test metric exceeds the threshold $Th_{i}$.

\begin{equation}
d_{\text{OOD}} = 
\begin{cases} 
0, & \text{if } s_1 \leq Th_1 \\ 
1, & \text{if } s_1 > Th_1 
\end{cases}
\end{equation}
The global OOD decision, $D_{\text{OOD}}$,  is derived through a voting-based aggregation of the local decisions and is given by
\begin{equation}
D_{\text{OOD}} = \prod_{i=1}^{C} d_{\text{OOD}}
\end{equation}
This approach implies that for a global OOD decision to indicate an OOD status, all local OOD decisions must indicate OOD (i.e., $d_{\text{OOD}}$ = 1 for all classes). To summarize, during the test phase, test metrics associated with label $i$ are sampled, and the mean of these test residuals is calculated. If any of the mean residual values for label $i$ exceed the predefined class threshold, the test sample is classified as belonging to a known target. If none of the mean residuals exceed their respective class thresholds, the test target is classified as unknown.

\begin{figure*}[htb!]
    \centering
    \begin{subfigure}[b]{0.45\textwidth}
        \includegraphics[width=\textwidth]{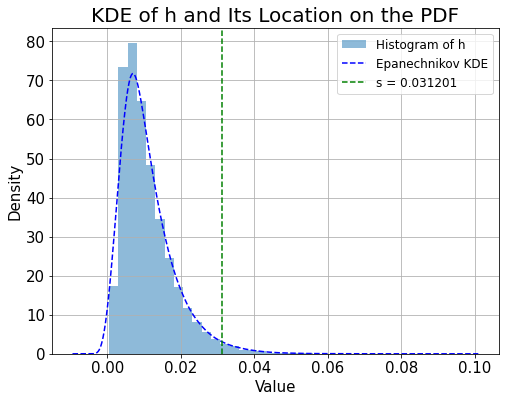}
        \caption{Training data histogram and its approximation using Epanechnikov Kernel Density Estimation}
        \label{fig:KDE and PDF}
    \end{subfigure}
    \hfill
    \begin{subfigure}[b]{0.45\textwidth}
        \includegraphics[width=\textwidth]{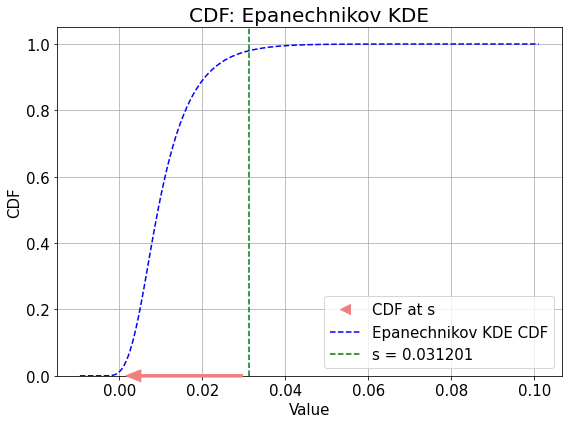}
        \caption{Approximated CDF and Percentile at s}
        \label{fig: CDF and s}
    \end{subfigure}
    \caption{Procedure to estimate the CDF of the training data distribution: \textit{(left)} Training data histogram and its approximation using Epanechnikov Kernel Density Estimation (KDE) \textit{(right)} Approximated CDF and percentile at s}
    \label{fig: Train Data Distribution Estimation}
\end{figure*}

\subsubsection{Global OOD Score}

AUROC and FPR95 calculations require a continuous scoring system. To achieve this, we transform our voting-based global OOD detection process into a probability-based representation. For each class-specific ID distribution $h$, an Epanechnikov Kernel Density Estimation (KDE) is applied, where $\hat{h}(x)$ denotes the estimated KDE of $h$, defined as:

\begin{equation}
\hat{h}(x) = \frac{1}{n \cdot B} \sum_{i=1}^{n} K\left( \frac{x - x_i}{B} \right)
\end{equation}

Here, $n$ represents the number of data points in the training distribution $h$, $B$ is the bandwidth controlling the smoothness, $x_{i}$ are the individual data points from the distribution $h$, and $K(u)$ is the kernel function, given by:
\begin{equation}
K(u) = \begin{cases} 
\frac{3}{4} (1 - u^2) & \text{if } |u| \leq 1, \\ 
0 & \text{otherwise}.
\end{cases}
\end{equation}

where u is the scaled variable given by $\frac{x - x_i}{B}$. This kernel density approximation enables the fitting shown in Figure 10(a), where the Epanechinikov KDE alignes well with the selected bandwidth $B$. Once $\hat{h}(x)$ is constructed, the CDF at the test-train residual's local OOD metric, $s_{i}$, can be estimated, as illustrated in Figure 10(b). The CDF value at $s_{i}$ in this example is located near the upper percentile of the distribution $h$, close to 1. Using the CDF directly, however, yields a biased probability of being OOD, as it does not account for the domain gap during the test phase. To address this, the CDF estimation is substituted into Equation (29), where $\beta$ is a scaling factor that adjusts the sensitivity of the OOD score. For our experiment, we set $\beta$ = 120, though this parameter could be optimized by examining the distributional behavior between in-class and inter-class residuals, a topic beyond the scope of this paper.

\begin{equation}
p_{\text{ood}}^{(i)} = \frac{1}{\exp\left(-\beta \cdot \left(1 - \text{CDF}_{\hat{h}_{i}}(s_i)\right)\right)}
\end{equation}

After calculating the $p_{ood}^{i}$, which represents the local OOD score as a probability, we compute the global OOD probability by taking the product of all class=specific probabilities: 
\begin{equation}
P_{\text{ood}}^{\text{global}} = \prod_{i=1}^{C} p_{\text{ood}}^{(i)}
\end{equation}
Using this global OOD probability, we can estimate AUROC and FPR95. This probability-based global OOD score can also replace the original voting-based aggregation method, which relied on localized thresholds, allowing for a single global OOD threshold with probabilistic interpretation. However, note that the resulting AUROC and FPR95 values may vary significantly depending on the choice of $\beta$; optimizing this parameter could further improve these metrics.

\subsection{Statistical Distance Method}

Table III and Table IV display the relative performance using 1-$\rho$ for feature-correlation based statistical distance with different aggregation methods for collected test samples with known labels, and 1-$\rho$ +$\sigma$ for feature-residual base statistical distance with varying aggregation methods, specifically highlighting the mean aggregation method for 1-$\rho$. The tables reveal that both approaches demonstrate comparable performance, with minimal accuracy variation across experiments and respective cases. Furthermore, the accuracy variation between known and unknown samples is relatively small in both methods, indicating stable ID/OOD prediction performance. This suggests that Pearson correlation between feature vectors and that between residual vectors provides similar information, as expected, given the analogy between the feature-correlation approach and the class-specific residual generation process based on Pearson correlation calculations.

Among the aggregation methods for test-train residual sampling, mean and min aggregation show better performance than the max method when data size is sufficient. Additionally, when the number of known targets exceeds unknown targets, mean and min aggregation methods yield higher overall accuracy than the max aggregation method. Conversely, when unknown targets outnumber known ones, the max method yields higher accuracy, consistent with statistical sampling theory.

\begin{table*}[htbp]
\centering
\caption{\textsc{Performance Comparison of Statistical Distance on Different BASES in Experiment 1}}
\begin{threeparttable}
\resizebox{1\textwidth}{!}{ 
\begin{tabular}{l ccc|cccc}
\hline
\hline
\multicolumn{8}{c}{Statistical Distance: Experiment 1} \\ \hline
Base & \multicolumn{3}{c}{Feature-Correlation Base} & \multicolumn{4}{c}{Feature-Residual Base} \\ 
Model & \(1 - \rho\) & \(1 - \rho\) & \(1 - \rho\) & \(1 - \rho + \sigma\) & \(1 - \rho + \sigma\) & \(1 - \rho + \sigma\) & \(1 - \rho\) \\ 

Aggregation & mean & max & min & mean & max & min & mean \\ \hline
Case 1: Accuracies$\uparrow$ & 87.38/85.51/89.10 & 83.75/95.11/73.32 & 87.38/73.05/95.23 & 87.52/88.16/86.82 & 82.35/70.52/95.23 & 85.58/95.23/75.07 & 87.15/85.21/88.93 \\ 
AUROC$\uparrow$/FPR95$\downarrow$ & 0.9208/0.2197 & - & - & 0.9164/0.2427 & - & - & 0.9211/0.2296 \\ 
Case 2: Accuracies$\uparrow$ & 86.86/88.79/85.10 & 81.55/96.06/68.22 & 86.86/76.39/92.71 & 87.46/82.85/92.49 & 79.07/62.03/97.61 & 86.69/91.67/81.28 & 86.86/88.55/85.32 \\ 
AUROC$\uparrow$/FPR95$\downarrow$ & 0.9035/0.2389 & - & - & 0.8958/0.2203 & - & - & 0.9046/0,2356 \\ 
Case 3: Accuracies$\uparrow$ & 76.78/79.31/74.47 & 74.01/88.49/60.71 & 76.78/68.63/83.84 & 77.73/74.14/81.63 & 74.36/59.07/91.00 & 76.84/83.40/69.71 & 76.93/79.31/74.74 \\ 
AUROC$\uparrow$/FPR95$\downarrow$ & 0.8069/0.5956 & - & - & 0.8149/0.5304 & - & - & 0.8082/0.5742 \\ 
Case 4: Accuracies$\uparrow$ & 67.39/58.02/76.00 & 66.73/71.79/62.08 & 67.39/39.42/88.71 & 68.62/79.12/57.19 & 69.10/67.45/70.90 & 66.53/88.82/42.28 & 67.16/58.14/75.45 \\ 
AUROC$\uparrow$/FPR95$\downarrow$ & 0.7259/0.7381 & - & - & 0.7435/0.7299 & - & - & 0.7239/0.7485 \\ 
Case 5: Accuracies$\uparrow$ & 51.83/20.45/80.66 & 52.34/38.34/65.21 & 51.83/8.29/92.82 & 53.63/95.40/8.17 & 53.83/94.08/10.02 & 53.48/95.84/7.39 & 52.08/20.81/80.82 \\ 
AUROC$\uparrow$/FPR95$\downarrow$ & 0.5324/0.9260 & - & - & 0.5738 & - & - & 0.5299/0.9282 \\ \hline

\end{tabular}
} 
\begin{tablenotes}
\footnotesize
\item Accuracies are reported in the following format: Overall Accuracy / Known Accuracy / Unknown Accuracy.
\item Aggregation method indicates how multiple residuals of test samples with known class labels are aggregated.
\end{tablenotes}
\end{threeparttable}
\end{table*}

\begin{table*}[htbp]
\centering
\caption{\textsc{Performance Comparison of Statistical Distance on Different BASES in Experiment 2}}
\begin{threeparttable}
\resizebox{1\textwidth}{!}{
\begin{tabular}{l ccc|cccc}
\hline
\hline
\multicolumn{8}{c}{Statistical Distance: Experiment 2} \\ \hline
Base & \multicolumn{3}{c}{Feature-Correlation Base} & \multicolumn{4}{c}{Feature-Residual Base} \\ 
Model & \(1 - \rho\) & \(1 - \rho\) & \(1 - \rho\) & \(1 - \rho + \sigma\) & \(1 - \rho + \sigma\) & \(1 - \rho + \sigma\) & \(1 - \rho\) \\ 

Aggregation & mean & max & min & mean & max & min & mean \\ \hline
Case 1: Accuracies$\uparrow$ & 76.78/79.31/74.47 & 74.01/88.49/60.71 & 76.78/68.63/83.84 & 77.73/74.14/81.63 & 74.36/59.07/91.00 & 76.84/83.40/69.71 & 76.93/79.31/74.74 \\ 
\phantom{Case 1: }AUROC$\uparrow$/FPR95$\downarrow$ & 0.8069/0.5956 & - & - & 0.8149/0.5304 & - & - & 0.8082/0.5742 \\ 
Case 2: Accuracies$\uparrow$ & 62.99/52.49/74.42 & 62.71/68.71/56.17 & 62.99/39.73/86.64 & 60.14/69.29/51.73 & 58.91/49.31/67.73 & 60.51/83.72/39.18 & 62.99/52.71/74.18 \\ 
\phantom{Case 1: }AUROC$\uparrow$/FPR95$\downarrow$ & 0.6559/0.9159 & - & - & 0.6205/0.9290 & - & - & 0.6555/0.9141 \\ 
Case 3: Accuracies$\uparrow$ & 74.93/66.02/82.46 & 75.79/82.92/69.76 & 74.93/45.07/90.89 & 76.19/82.67/68.52 & 75.90/68.39/84.79 & 69.93/92.26/43.52 & 75.30/66.83/82.46 \\ 
\phantom{Case 1: }AUROC$\uparrow$/FPR95$\downarrow$ & 0.8275/0.5369 & - & - & 0.8361/0.4858 & - & - & 0.8293/0.5074 \\ 
Case 4: Accuracies$\uparrow$ & 81.84/83.17/81.00 & 72.84/95.72/58.41 & 81.84/66.72/93.20 & 82.24/81.51/83.39 & 72.10/57.76/94.83 & 82.32/93.57/64.50 & 81.95/83.39/81.04 \\ 
\phantom{Case 1: }AUROC$\uparrow$/FPR95$\downarrow$ & 0.8682/0.3898 & - & - & 0.8661/0.3801 & - & - & 0.8667/0.3982 \\ 
Case 5: Accuracies$\uparrow$ & 77.36/65.33/94.70 & 81.67/77.08/88.28 & 77.36/51.98/97.14 & 80.87/94.91/71.13 & 86.09/88.08/84.72 & 67.56/97.56/46.76 & 77.56/65.72/94.63 \\ 
\phantom{Case 1: }AUROC$\uparrow$/FPR95$\downarrow$ & 0.9103/0.3898 & - & - & 0.9286/0.3361 & - & - & 0.9100/0.3961 \\ \hline

\end{tabular}
}
\begin{tablenotes}
\footnotesize
\item Accuracies are reported in the following format: Overall Accuracy / Known Accuracy / Unknown Accuracy.
\item Aggregation method indicates how multiple residuals of test samples with known class labels are aggregated.
\end{tablenotes}

\end{threeparttable}
\end{table*}

\subsection{OpenMax}

Table V compares the relative performance of the OpenMax OOD detection method between feature-based and residual-based approaches. Overall accuracies and AUROC scores generally show significant improvement. Additionally, the gap between known and unknown accuracies narrows in most cases, indicating improved ID/OOD stability, except in Experiment 1, Case 5, where the training dataset is limited. This consistent improvement is particularly notable given that the primary OpenMax algorithm relies on a Weibull distribution model. These findings suggest that class-specific residual transformation provides a more robust and interpretable reference space, better suited to statistical model fitting.

\begin{table*}[htbp]
\centering
\caption{\textsc{Performance Comparison of OpenMax on Different bases}}
\begin{threeparttable}
\begin{tabular}{l cc cc}
\hline
\hline
\multicolumn{5}{c}{OpenMax} \\ \hline
& \multicolumn{2}{c}{Experiment 1} & \multicolumn{2}{c}{Experiment 2} \\ 
Base & \multicolumn{1}{c}{Feature Base} & \multicolumn{1}{c}{Residual Base} & \multicolumn{1}{c}{Feature Base} & \multicolumn{1}{c}{Residual Base} \\ 
Model & OpenMax & OpenMax & OpenMax & OpenMax \\ 
Aggregation & - & mean & - & mean \\ \hline
Case 1: Accuracies$\uparrow$ & 78.41 / 62.25 / 96.00 & 87.46 / 89.42 / 85.33 & 63.79 / 32.66 / 96.67 & 77.18 / 71.89 / 82.95 \\ 
\phantom{Case 1: }AUROC$\uparrow$/FPR95$\downarrow$ & 0.8469 / 0.3595 & 0.9133 / 0.4203 & 0.7419 / 0.5945 & 0.8305 / 0.5589 \\ 
Case 2: Accuracies$\uparrow$ & 69.76 / 43.29 / 98.57 & 85.55 / 80.60 / 90.94 & 54.17 / 21.05 / 84.60 & 60.14 / 67.20 / 53.64 \\ 
\phantom{Case 2: }AUROC$\uparrow$/FPR95$\downarrow$ & 0.8146 / 0.4537 & 0.9066 / 0.2559 & 0.6152 / 0.9278 & 0.6501 / 0.8706 \\ 
Case 3: Accuracies$\uparrow$ & 63.79 / 32.66 / 97.67 & 77.18 / 71.89 / 82.95 & 67.22 / 43.47 / 95.32 & 74.44 / 78.61 / 69.51 \\ 
\phantom{Case 3: }AUROC$\uparrow$/FPR95$\downarrow$ & 0.7419 / 0.5945 & 0.8305 / 0.5589 & 0.6360 / 0.6154 & 0.8320 / 0.5548 \\ 
Case 4: Accuracies$\uparrow$ & 59.65 / 28.33 / 93.74 & 67.96 / 70.41 / 65.30 & 68.08 / 50.12 / 96.53 & 80.04 / 84.44 / 73.06 \\ 
\phantom{Case 4: }AUROC$\uparrow$/FPR95$\downarrow$ & 0.7413 / 0.7578 & 0.7491 / 0.7847 & 0.6152 / 0.5435 & 0.8710 / 0.4364 \\ 
Case 5: Accuracies$\uparrow$ & 52.51 / 44.93 / 60.76 & 53.06 / 75.01 / 29.16 & 87.44 / 85.63 / 88.68 & 80.07 / 95.61 / 69.29 \\ 
\phantom{Case 5: }AUROC$\uparrow$/FPR95$\downarrow$ & 0.5505 / 0.9255 & 0.5931 / 0.9299 & 0.8817 / 0.3389 & 0.9067 / 0.6876 \\ \hline

\end{tabular}
\begin{tablenotes}
\footnotesize
\item Accuracies are reported in the following format: Overall Accuracy / Known Accuracy / Unknown Accuracy.
\item Aggregation method indicates how multiple residuals of test samples with known class labels are aggregated.
\end{tablenotes}
\end{threeparttable}
\end{table*}

\subsection{Mahalanobis Distance}

Table VI presents the relative performance of the Mahalanobis distance OOD detection method when comparing feature-based and residual-based approaches. Overall, the residual-based approach shows moderate accuracy improvements across different training dataset sizes and varying numbers and types of unknown targets, with a slightly narrower accuracy gap between known and unknown targets. However, as observed in Experiment 1, AUROC tends to decline as data size decreases with the residual-based transformation. In contrast, Experiment 2, which maintains a consistent training size but varies the composition of known and unknown targets, generally shows an upward trend in AUROC.
The reduced AUROC observed in Experiment 1 with limited training samples could be due to the constrained sampling of test-train residual combinations, implemented to reduce computational costs. For each local OOD score calculation, CDF values at $s_{i}$ could more accurately approximate the distribution with an increased number of test-train residual samples. However, in Experiment 1, reduced test-train sampling and smaller training sets may not yield a CDF representative of the true distribution. Further experiments with expanded sampling of test-train residuals are needed to confirm this effect. Additionally, to enhance scalability in real-world applications without excessive computational costs, modified algorithms that selectively use inter-class residuals alongside feature residuals could potentially mitigate the challenges associated with smaller training datasets.

\begin{table*}[htbp]
\centering
\caption{\textsc{Performance Comparison of Mahalanobis Distance Model on Different Bases}}
\begin{threeparttable}
\begin{tabular}{l cc cc}
\hline
\hline
\multicolumn{5}{c}{Mahalanobis Distance} \\ \hline
& \multicolumn{2}{c}{Experiment 1} & \multicolumn{2}{c}{Experiment 2} \\ 
Base & \multicolumn{1}{c}{Feature Base} & \multicolumn{1}{c}{Residual Base} & \multicolumn{1}{c}{Feature Base} & \multicolumn{1}{c}{Residual Base} \\ 
Model & MD & MD & MD & MD \\ 
Aggregation & - & mean & - & mean \\ \hline
Case 1: Accuracies$\uparrow$ & 55.60 / 98.74 / 8.65 & 59.88 / 99.07 / 17.23 & 82.58 / 85.32 / 79.61 & 88.29 / 87.78 / 88.85 \\ 
\phantom{Case 1: }AUROC$\uparrow$/FPR95$\downarrow$ & 0.9290 / 0.3441 & 0.9700 / 0.12 & 0.9100 / 0.3162 & 0.8694 / 0.2471 \\ 
Case 2: Accuracies$\uparrow$ & 63.51 / 98.36 / 25.58 & 74.64 / 98.30 / 48.90 & 45.77 / 85.93 / 8.88 & 58.37 / 83.42 / 35.34 \\ 
\phantom{Case 2: }AUROC$\uparrow$/FPR95$\downarrow$ & 0.9224 / 0.3573 & 0.9661 / 0.1063 & 0.4492 / 0.9410 & 0.6309 / 0.8646 \\ 
Case 3: Accuracies$\uparrow$ & 82.58 / 85.32 / 79.61 & 88.29 / 87.78 / 88.85 & 66.96 / 89.99 / 39.71 & 63.59 / 91.99 / 29.99 \\ 
\phantom{Case 3: }AUROC$\uparrow$/FPR95$\downarrow$ & 0.9100 / 0.3162 & 0.8694 / 0.2471 & 0.7158 / 0.8541 & 0.8109 / 0.5227 \\ 
Case 4: Accuracies$\uparrow$ & 48.89 / 1.92 / 100 & 49.31 / 2.74 / 100 & 61.82 / 100 / 1.33 & 62.02 / 99.95 / 1.92 \\ 
\phantom{Case 4: }AUROC$\uparrow$/FPR95$\downarrow$ & 0.8488 / 0.4822 & 0.8425 / 0.5474 & 0.4626 / 0.9078 & 0.8711 / 0.4392 \\ 
Case 5: Accuracies$\uparrow$ & 48.23 / 0.66 / 100 & 53.46 / 96.82 / 6.26 & 53.28 / 99.02 / 21.57 & 62.88 / 98.95 / 37.86 \\ 
\phantom{Case 5: }AUROC$\uparrow$/FPR95$\downarrow$ & 0.7986 / 0.6164 & 0.7019 / 0.8044 & 0.9184 / 0.3815 & 0.9438 / 0.2392 \\ \hline

\end{tabular}
\begin{tablenotes}
\footnotesize
\item MD: Mahalanobis Distance
\item Accuracies are reported in the following format: Overall Accuracy / Known Accuracy / Unknown Accuracy.
\item Aggregation method indicates how multiple residuals of test samples with known class labels are aggregated.
\end{tablenotes}
\end{threeparttable}
\end{table*}

\subsection{ViM}
Tables VII and VIII compare the relative performance of ViM between feature-based and residual-based approaches. The ViM method appears less suitable for the residual-based approach without further customization, as results in Table VII do not fully support improved OOD detection accuracy in SAR imagery. ViM decomposes feature vectors into principal components and their residuals, inherently incorporating residual processing within the ID reference space. As expected, trends in overall accuracy, AUROC, and FPR95 remain inconclusive. In some cases, the overall accuracy with a predefined threshold setting may even decline when transformed to the residual-based approach. This outcome may result from decomposition-based feature processing, which disrupts the distributional patterns of residual inputs. Thus, ViM's underlying OOD detection algorithm is not compatible with distributional input-based methods.

\begin{table*}[htbp]
\centering
\caption{\textsc{Performance Comparison of ViM on Different Bases in Experiment 1}}
\begin{threeparttable}
\begin{tabular}{l cc cc}
\hline
\hline
\multicolumn{5}{c}{ViM: Experiment 1} \\ \hline
Base & \multicolumn{2}{c}{Feature Base} & \multicolumn{2}{c}{Residual Base} \\ 
Model & ViM & ViM & ViM & ViM \\

Overwrite? & $\checkmark$ & \ding{55} & $\checkmark$ & \ding{55} \\ \hline
Case 1: Accuracies$\uparrow$ & 63.08 / 95.56 / 27.73 & 66.59 / 89.86 / 41.26 & 63.39 / 62.19 / 64.70 & 56.82/21.04/95.77 \\ 
\phantom{Case 1: }AUROC$\uparrow$/FPR95$\downarrow$ & 0.7895 / 0.6740 & 0.7926 / 0.6553 & 0.6976 / 0.7430 & 0.8374 / 0.5107 \\ 
Case 2: Accuracies$\uparrow$ & 54.77 / 98.90 / 6.74 & 59.97 / 95.51 / 21.29 & 58.77 / 52.00 / 66.13 & 57.85 / 23.23 / 95.53 \\ 
\phantom{Case 1: }AUROC$\uparrow$/FPR95$\downarrow$ & 0.7209 / 0.7863 & 0.7255 / 0.7918 & 0.7243 / 0.8756 & 0.7992 / 0.7501 \\ 
Case 3: Accuracies$\uparrow$ & 67.93 / 90.58 / 43.29 & 69.39 / 80.88 / 56.89 & 58.77 / 79.89 / 35.78 & 56.77 / 79.89 / 35.78 \\ 
\phantom{Case 1: }AUROC$\uparrow$/FPR95$\downarrow$ & 0.7494 / 0.8849 & 0.7453 / 0.8860 & 0.7706 / 0.7386 & 0.8111 / 0.7578 \\ 
Case 4: Accuracies$\uparrow$ & 68.42 / 80.71 / 55.04 & 67.13 / 69.59 / 64.46 & 67.76 / 85.53 / 48.42 & 55.88 / 22.85 / 91.83 \\ 
\phantom{Case 1: }AUROC$\uparrow$/FPR95$\downarrow$ & 0.7280 / 0.9189 & 0.7225 / 0.9211 & 0.7747 / 0.7063 & 0.8081 / 0.6822 \\ 
Case 5: Accuracies$\uparrow$ & 52.37 / 100 / 0.54 & 53.51 / 99.78 / 3.16 & 49.20 / 83.89 / 11.45 & 48.14 / 49.53 / 40.63 \\ 
\phantom{Case 1: }AUROC$\uparrow$/FPR95$\downarrow$ & 0.5456 / 0.9803 & 0.5458 / 0.9803 & 0.7065 / 0.8241 & 0.6999 / 0.8515 \\ \hline

\end{tabular}
\begin{tablenotes}
\footnotesize
\item Accuracies are reported in the following format: Overall Accuracy / Known Accuracy / Unknown Accuracy.
\item Overwrite? indicates whether, when ViM scores are appended to the logits list, the original unknown class logit value is overwritten (and thus excluded from the calculation of the softmax-converted ViM score) or retained (and included in the calculation of the softmax-converted ViM score).

\end{tablenotes}
\end{threeparttable}
\end{table*}

\begin{table*}[htbp]
\centering
\caption{\textsc{Performance Comparison of ViM on Different Bases in Experiment 2}}
\begin{threeparttable}
\begin{tabular}{l cc cc}
\hline
\hline
\multicolumn{5}{c}{ViM: Experiment 2} \\ \hline
Base & \multicolumn{2}{c}{Feature Base} & \multicolumn{2}{c}{Residual Base} \\ 
Model & ViM & ViM & ViM & ViM \\

Overwrite? & $\checkmark$ & \ding{55} & $\checkmark$ & \ding{55} \\ \hline
Case 1: Accuracies$\uparrow$ & 67.93 / 90.58 / 43.29 & 69.39 / 80.88 / 56.89 & 58.77 / 79.89 / 35.78 & 56.77 / 79.89 / 35.78 \\ 
\phantom{Case 1: }AUROC$\uparrow$/FPR95$\downarrow$ & 0.7494 / 0.8849 & 0.7453 / 0.8860 & 0.7706 / 0.7386 & 0.8111 / 0.7578 \\ 
Case 2: Accuracies$\uparrow$ & 48.29 / 99.64 / 1.10 & 47.66 / 92.02 / 2.30 & 35.04 / 42.04 / 28.06 & 50.37 / 15.80 / 82.14 \\ 
\phantom{Case 1: }AUROC$\uparrow$/FPR95$\downarrow$ & 0.5056 / 0.9678 & 0.5226 / 0.9024 & 0.3077 / 0.9726 & 0.3118 / 0.9720 \\ 
Case 3: Accuracies$\uparrow$ & 55.68 / 99.63 / 3.68 & 58.82 / 94.52 / 16.58 & 52.74 / 89.04 / 9.79 & 42.09 / 20.92 / 67.14 \\ 
\phantom{Case 1: }AUROC$\uparrow$/FPR95$\downarrow$ & 0.7156 / 0.7561 & 0.7207 / 0.7434 & 0.6215 / 0.9484 & 0.6744 / 0.8430 \\ 
Case 4: Accuracies$\uparrow$ & 56.11 / 78.57 / 20.52 & 56.31 / 72.43 / 30.77 & 52.83 / 56.92 / 46.35 & 35.64 / 8.24 / 79.04 \\ 
\phantom{Case 1: }AUROC$\uparrow$/FPR95$\downarrow$ & 0.5906 / 0.7885 & 0.5973 / 0.7783 & 0.3920 / 0.9544 & 0.4840 / 0.8696 \\ 
Case 5: Accuracies$\uparrow$ & 40.95 / 100 / 0 & 41.66 / 100 / 1.21 & 47.00 / 84.80 / 20.79 & 56.28 / 19.11 / 82.06 \\ 
\phantom{Case 1: }AUROC$\uparrow$/FPR95$\downarrow$ & 0.8574 / 0.5635 & 0.8727 / 0.5356 & 0.7688 / 0.7615 & 0.8521 / 0.5307 \\ \hline

\end{tabular}
\begin{tablenotes}
\footnotesize
\item Accuracies are reported in the following format: Overall Accuracy / Known Accuracy / Unknown Accuracy.
\item Overwrite? indicates whether, when ViM scores are appended to the logits list, the original unknown class logit value is overwritten (and thus excluded from the calculation of the softmax-converted ViM score) or retained (and included in the calculation of the softmax-converted ViM score).

\end{tablenotes}

\end{threeparttable}
\end{table*}

\subsection{kNN Algorithms}
Tables IX and X compare the performance of kNN-based OOD detection methods using distance-based and density-based metrics across feature-based and residual-based approaches. The distance-based kNN algorithm determines OOD status by assessing the proximity between a test sample and its nearest neighbors in feature space, flagging a sample as OOD if its distance from the k-nearest neighbors exceeds a defined threshold. In the density-based kNN approach, a 1/distance metric is employed; while this inversion maintains accuracy by adjusting the threshold accordingly, AUROC values can vary due to differences in the CDF approximation process.

In Experiment 1 (excluding Case 4) and Experiment 2 (excluding Case 5), overall performance shows marginal improvement, though AUROC generally decreases across both experiments. This suggests that simple Euclidean distance-based separations, as applied to distributional inputs with feature-residual transformations, may not be suitable to enhance both accuracy and AUROC. As distance (or density) metrics primarily reflect isolated, individual pattern characteristics rather than cohesive, group-level structures, these findings indicate the need for transitioning from distance-based metrics to permutation-invariant metrics to improve OOD detection effectiveness.
\begin{table*}[htbp]
\centering
\caption{\textsc{Performance Comparison of kNN algorithms on Different Bases in Experiment 1}}
\begin{threeparttable}
\begin{tabular}{l c cc}
\hline
\hline
\multicolumn{4}{c}{kNN Algorithms: Experiment 1} \\ \hline
Base & \multicolumn{1}{c}{Feature Base} & \multicolumn{2}{c}{Residual Base} \\ 
Model & kNN & kNN & kNN \\

Metrics & Distance / Density & Distance & Density \\ \hline
Case 1: Accuracies$\uparrow$ & 81.90 / 69.10 / 95.83 & \multicolumn{2}{c}{87.49 / 87.89 / 87.06} \\ 
\phantom{Case 1: }AUROC$\uparrow$/FPR95$\downarrow$ & 0.9455 / 0.2773 & 0.8657 / 0.2866 & 0.9098 / 0.2877 \\ 
Case 2: Accuracies$\uparrow$ & 76.16 / 56.22 / 97.85 & \multicolumn{2}{c}{86.38 / 81.10 / 92.13} \\ 
\phantom{Case 2: }AUROC$\uparrow$/FPR95$\downarrow$ & 0.9414 / 0.2373 & 0.8835 / 0.2553 & 0.8862 / 0.2553 \\ 
Case 3: Accuracies$\uparrow$ & 76.04 / 63.18 / 90.04 & \multicolumn{2}{c}{81.04 / 81.48 / 80.56} \\ 
\phantom{Case 3: }AUROC$\uparrow$/FPR95$\downarrow$ & 0.8778 / 0.5518 & 0.8493 / 0.4915 & 0.8500 / 0.4899 \\ 
Case 4: Accuracies$\uparrow$ & 71.56 / 70.90 / 72.27 & \multicolumn{2}{c}{65.73 / 92.33 / 36.79} \\ 
\phantom{Case 4: }AUROC$\uparrow$/FPR95$\downarrow$ & 0.7915 / 0.7315 & 0.7482 / 0.6871 & 0.7928 / 0.6866 \\ 
Case 5: Accuracies$\uparrow$ & 52.80 / 99.12 / 2.39 & \multicolumn{2}{c}{53.46 / 98.41 / 4.53} \\ 
\phantom{Case 5: }AUROC$\uparrow$/FPR95$\downarrow$ & 0.5959 / 0.9118 & 0.6220 / 0.8751 & 0.6248 / 0.8740 \\ \hline

\end{tabular}
\begin{tablenotes}
\footnotesize
\item Accuracies are reported in the following format: Overall Accuracy / Known Accuracy / Unknown Accuracy.
\item Aggregation method indicates how multiple residuals of test samples with known class labels are aggregated.
\end{tablenotes}
\end{threeparttable}
\end{table*}

\begin{table*}[htbp]
\centering
\caption{\textsc{Performance Comparison of kNN algorithms on Different Bases in Experiment 2}}
\begin{threeparttable}
\begin{tabular}{l c cc}
\hline
\hline
\multicolumn{4}{c}{kNN Algorithms: Experiment 2} \\ \hline
Base & \multicolumn{1}{c}{Feature Base} & \multicolumn{2}{c}{Residual Base} \\ 
Model & kNN & kNN & kNN \\

Metrics & Distance / Density & Distance & Density \\ \hline
Case 1: Accuracies$\uparrow$ & 76.04 / 63.18 / 90.04 & \multicolumn{2}{c}{81.04 / 81.48 / 80.56} \\ 
\phantom{Case 1: }AUROC$\uparrow$/FPR95$\downarrow$ & 0.8778 / 0.5518 & 0.8493 / 0.4915 & 0.8500 / 0.4899 \\ 
Case 2: Accuracies$\uparrow$ & 56.48 / 40.85 / 70.85 & \multicolumn{2}{c}{59.59 / 70.78 / 49.32} \\ 
\phantom{Case 2: }AUROC$\uparrow$/FPR95$\downarrow$ & 0.6306 / 0.9547 & 0.5939 / 0.9457 & 0.5977 / 0.9457 \\ 
Case 3: Accuracies$\uparrow$ & 74.64 / 62.91 / 88.53 & \multicolumn{2}{c}{75.07 / 82.24 / 66.58} \\ 
\phantom{Case 3: }AUROC$\uparrow$/FPR95$\downarrow$ & 0.8403 / 0.4868 & 0.7983 / 0.4900 & 0.8286 / 0.4905 \\ 
Case 4: Accuracies$\uparrow$ & 70.30 / 53.89 / 96.31 & \multicolumn{2}{c}{82.04 / 80.58 / 84.35} \\ 
\phantom{Case 4: }AUROC$\uparrow$/FPR95$\downarrow$ & 0.9108 / 0.3987 & 0.8564 / 0.4360 & 0.8574 / 0.4378 \\ 
Case 5: Accuracies$\uparrow$ & 86.84 / 88.08 / 88.98 & \multicolumn{2}{c}{79.61 / 96.44 / 67.94} \\ 
\phantom{Case 5: }AUROC$\uparrow$/FPR95$\downarrow$ & 0.9374 / 0.3319 & 0.8916 / 0.3082 & 0.9368 / 0.3054 \\ \hline

\end{tabular}
\begin{tablenotes}
\footnotesize
\item Accuracies are reported in the following format: Overall Accuracy / Known Accuracy / Unknown Accuracy.
\item Aggregation method indicates how multiple residuals of test samples with known class labels are aggregated.
\end{tablenotes}
\end{threeparttable}
\end{table*}


\section{Discussion}

Among the feature-based OOD detection methods tested, OpenMax and Mahalanobis distance methods demonstrated promising results. Both methods rely on an assumed statistical distribution of inputs, aligning effectively with our statistical formalism. In contrast, ViM and kNN algorithms showed either negligible or adverse effects on performance metrics, indicating their limited suitability for residual-based transformations. This result is expected, as ViM inherently incorporates a residual decomposition concept, and kNN is known to underperform when intrinsic feature information is reduced. Additionally, kNN’s OOD detection relies on individual component patterns rather than distributional patterns, which further limits its adaptability to residual-based inputs. Residuals are primarily designed to capture subtle deviations from normal distributions for OOD detection, rather than depending on intrinsic feature characteristics.

A notable observation from the tables is the comparison between known and unknown target detection accuracy. Our proposed method consistently demonstrates a narrower discrepancy between known and unknown detection accuracy, indicating stability in detection rates independent of overall accuracy. This consistency supports more effective threshold decision-making, as both accuracy metrics exhibit reduced susceptibility to large fluctuations. Although our method generally enhances overall accuracy, performance declines as the training dataset size decreases. This may be an effect of reduced train-test sampling, especially when limited training data does not approximate the true population distribution. This limitation underscores the need for further estimation techniques to determine an optimal test-train residual sampling size.

If increasing test-train residual samples is not computationally feasible, further adaptation of OOD detection methods compatible with distributional inputs could be considered. Although only in-class residuals were used in this study, both in-class and inter-class residuals could be incorporated by assigning binary labels of 1 and 0, respectively. In this framework, standardized residuals could be interpreted as distributional inputs with binary labels, making permutation-invariant deep set approaches \cite{b62, b72} or 2-Wasserstein distance-based optimal transport algorithms promising candidates in line with our findings \cite{b63, b68, b69, b70, b71}.

\section{Conclusion}
We propose a novel approach that transforms the feature space into a class-localized residual space, significantly enhancing OOD detection performance in SAR images. Conventional deep learning-based neural networks often struggle to establish resilient reference feature spaces capable of differentiating a diverse range of unknown targets. Our approach addresses this challenge, which is especially prevalent in SAR image-based unknown target detection, where decision boundaries frequently rely on predefined thresholds and inadequately constructed reference spaces.

Our residual-based approach enhances traditional feature-based OOD algorithms by amplifying intrinsic signals critical for anomaly detection while effectively reducing clutter and noise. This transformation constructs a more robust reference space, converting feature vectors into standardized residuals that are more interpretable as probability distributions. Additionally, by redefining the feature space, our class-specific residual approach is compatible with other feature-based OOD detection algorithms that utilize distributional inputs. We demonstrate that these reshaped residual distributions improve the detection of unclassified targets, often outperforming feature-based anomaly detection methods across diverse known/unknown training conditions. This shift to a residual-based framework remarks a significant milestone, especially for applications involving noisy or low-signal data, such as SAR imagery. Future work will focus on developing additional mathematical frameworks to capture the subtle patterns within the distribution of feature residuals, further advancing this approach.


%

\appendices

\section*{Acknowledgment}
The authors thank the Pohang University of Science and Technology for its generous support.

\ifCLASSOPTIONcaptionsoff
  \newpage
\fi

\end{document}